\documentclass[sigconf]{acmart}

\usepackage{multirow}
\usepackage[x11names,table]{xcolor}
\usepackage{mathtools}
\usepackage{subcaption}
\usepackage{algorithm,algorithmic}
\usepackage[nolist]{acronym}
\usepackage{hyperref}
\AtBeginDocument{%
  }

\setcopyright{cc}
\copyrightyear{2025} 
\acmYear{2025}
\setcctype{by}
\acmConference[FAccT '25]{The 2025 ACM Conference on Fairness, Accountability, and Transparency}{June 23--26, 2025}{Athens, Greece}
\acmBooktitle{The 2025 ACM Conference on Fairness, Accountability, and Transparency (FAccT '25), June 23--26, 2025, Athens, Greece}\acmDOI{10.1145/3715275.3732151}
\acmISBN{979-8-4007-1482-5/2025/06}
\begin{document}

\begin{acronym}
\acro{nn}[NN]{neural network}
\acro{xai}[XAI]{eXplainable Artificial Intelligence}
\end{acronym}

\title{From Confusion to Clarity: ProtoScore - A Framework for Evaluating Prototype-Based XAI}
\author{Helena Monke}
\email{helena.monke@ipa.fraunhofer.de}
\orcid{0009-0002-9844-4277}
\affiliation{\institution{Fraunhofer Institute for Manufacturing Engineering and Automation IPA}
  \city{Stuttgart}
  \country{Germany},\\ 
  \institution{Institute of Industrial Manufacturing and Management IFF, University of Stuttgart}
  \city{Stuttgart}
  \country{Germany}
}

\author{Benjamin Sae-Chew}
\email{benjamin.sae-chew@ipa.fraunhofer.de}
\orcid{0009-0009-2559-3950}
\affiliation{%
  \institution{Fraunhofer Institute for Manufacturing Engineering and Automation IPA}
  \city{Stuttgart}
  \country{Germany},\\
  \institution{TUM School of Computation, Information and Technology, Technical University of Munich}
  \city{Munich}
  \country{Germany}
}

\author{Benjamin Fresz}
\email{benjamin.fresz@ipa.fraunhofer.de}
\orcid{0009-0002-7463-8907}
\affiliation{%
  \institution{Fraunhofer Institute for Manufacturing Engineering and Automation IPA}
  \city{Stuttgart}
  \country{Germany},\\
  \institution{Institute of Industrial Manufacturing and Management IFF, University of Stuttgart}
  \city{Stuttgart}
  \country{Germany}
}

\author{Marco F. Huber}
\email{marco.huber@ieee.org}
\affiliation{\institution{Fraunhofer Institute for Manufacturing Engineering and Automation IPA}
  \city{Stuttgart}
  \country{Germany},\\
  \institution{Institute of Industrial Manufacturing and Management IFF, University of Stuttgart}
  \city{Stuttgart}
  \country{Germany}
  }

%\renewcommand{\shortauthors}{Monke, et al.}

\begin{abstract}
The complexity and opacity of \acp{nn} pose significant challenges, particularly in high-stakes fields such as healthcare, finance, and law, where understanding decision-making processes is crucial. To address these issues, the field of \ac{xai} has developed various methods aimed at clarifying AI decision-making, thereby facilitating appropriate trust and validating the fairness of outcomes. Among these methods, prototype-based explanations offer a promising approach that uses representative examples to elucidate model behavior. However, a critical gap exists regarding standardized benchmarks to objectively compare prototype-based \ac{xai} methods, especially in the context of time series data. This lack of reliable benchmarks results in subjective evaluations, hindering progress in the field. We aim to establish a robust framework, ProtoScore, for assessing prototype-based \ac{xai} methods across different data types with a focus on time series data, facilitating fair and comprehensive evaluations. By integrating the Co-12 properties of Nauta et al., this framework allows for effectively comparing prototype methods against each other and against other \ac{xai} methods, ultimately assisting practitioners in selecting appropriate explanation methods while minimizing the costs associated with user studies. All code is publicly available via \href{https://github.com/HelenaM23/ProtoScore}{GitHub}.%This ensures the effectiveness of explanations in real-world scenarios.
\end{abstract}

\begin{CCSXML}
<ccs2012>
<concept>
<concept_id>10010147.10010257</concept_id>
<concept_desc>Computing methodologies~Machine learning</concept_desc>
<concept_significance>500</concept_significance>
</concept>
<concept>
<concept_id>10002944.10011123.10011130</concept_id>
<concept_desc>General and reference~Evaluation</concept_desc>
<concept_significance>500</concept_significance>
</concept>
<concept>
<concept_id>10002944.10011123.10011124</concept_id>
<concept_desc>General and reference~Metrics</concept_desc>
<concept_significance>500</concept_significance>
</concept>
<concept>
<concept_id>10003120.10003121</concept_id>
<concept_desc>Human-centered computing~Human computer interaction (HCI)</concept_desc>
<concept_significance>300</concept_significance>
</concept>
</ccs2012>
\end{CCSXML}

\ccsdesc[500]{Computing methodologies~Machine learning}
\ccsdesc[500]{General and reference~Evaluation}
\ccsdesc[500]{General and reference~Metrics}
\ccsdesc[300]{Human-centered computing~Human computer interaction (HCI)}

\keywords{eXplainable AI, XAI, XAI properties, prototype explanation, explanation metrics, benchmark, technical evaluation, objective evaluation, automated evaluation}
 % \Description{Enjoying the baseball game from the third-base
 % seats. Ichiro Suzuki preparing to bat.}
 % \label{fig:teaser}

%\received{22 January 2025}

\maketitle

\section{Introduction}
% \begin{figure}
%     \centering
%     \includegraphics[trim=0cm 0.5cm 0cm 1cm, clip,width=1.03\linewidth]{prototype_architecture.pdf}
%     \caption{Architecture of a typical prototype network~\cite{Li_Liu_Chen_Rudin_2018}. It consists of an autoencoder and a classification network. In the prototype layer, the data samples are represented, e.g., as weighted distances of their latent representations to the prototypes. Other prototype methods may have slightly different structures. The benchmark focuses primarily on the latent space marked with the red rectangular.}
%     \label{fig:prototypearchitecture}
% \end{figure}

\begin{figure}
    \includegraphics[trim=6.5cm 18.5cm 6.5cm 2.8cm, clip, width=1\linewidth]{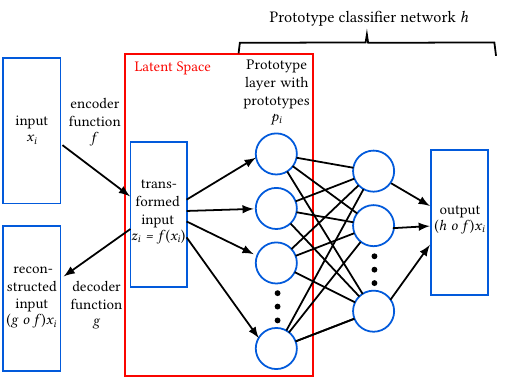}
    \caption{Architecture of a typical prototype network following~\cite{Li_Liu_Chen_Rudin_2018}. It consists of an autoencoder and a classification network. In the prototype layer, the data samples are represented, e.g., as weighted distances of their latent representations to the prototypes. Other prototype methods may have slightly different structures. The benchmark focuses primarily on the latent space marked with the red rectangular.}
    \label{fig:prototypearchitecture}
\end{figure}
\Acfp{nn} have become essential tools in machine learning, showing strong performance in applications such as image recognition, natural language processing, and many others. However, their high complexity and lack of transparency present challenges that limit their widespread use. The {\textquotedblleft black-box\textquotedblright} nature of \acp{nn} complicates users ability to understand how decisions are made and the reasons behind the model's behaviors~\cite{Rudin19}. This is especially relevant in critical fields such as healthcare, finance, and law, where decisions can have significant consequences~\cite{dzindolet2003}. 
The importance of explanations for AI systems has also been acknowledged by policymakers and courts as shown by XAI methods being used in court cases such as in the Australian case of ACCC v Trivago \cite{Fraser22}, or rulings on the use of XAI by the Court of Justice of the European Union \cite{CJEU2023, CJEU2025}, and the—now explicit—{\textquotedblleft Right to Explanation\textquotedblright} within the GDPR \cite{EuropeanParliament2016a} and AI Act Art. 86 \cite{EuropeanParliament2024}.
To tackle the aforementioned issues, the field of \acf{xai} has developed methods to clarify AI decision-making, enhancing human understanding of AI outputs and validating the fairness and reliability of those decisions. \Ac{xai} techniques offer insights into a model's reasoning, aiding in identifying biases or flaws, and thus calibrate user trust~\cite{McGuirl2006SupportingTC}. In addition, there exist several approaches to guide practitioners in selecting the most appropriate \ac{xai} methods for the task at hand~\cite{Theissler2022ExplainableAF, Tripathy.2022}.

One class of \ac{xai} methods uses {\textquotedblleft prototypes\textquotedblright}, which are representative examples that elucidate a model's behavior. A typical architecture of a prototype network is shown in Figure~\ref{fig:prototypearchitecture}. The encoder maps the input data to a latent space, where prototypes are computed for data subgroups in the prototype layer, enabling classification based on proximity to these reference points. The decoder transforms prototypes back to the input space. Prototype-based methods are inherently interpretable and generate predictions and explanations by comparing input data to learned prototypes representing the training data~\cite{narayanan2024prototypebased}. These prototypes, which closely resemble the training observations, collectively represent the entire dataset~\cite{Li_Liu_Chen_Rudin_2018}. They effectively map class examples to outputs, offering clarity through similarity reasoning and offering intuitive insights into the model's decision-making process. This approach emphasizes representative examples rather than relying on post-hoc explanations based on individual feature contributions (so-called {\textquotedblleft Saliency Maps\textquotedblright}), where \emph{post-hoc} describes explanations being created after model training and inference, in contrast to \emph{ante-hoc} explanations, which are part of the model itself. Additionally, visualizing prototypes with input data helps identifying patterns and anomalies, enhancing comprehension of
the underlying data distribution and model behavior.

For prototype methods, a major challenge remains: the lack of standardized benchmarks to objectively compare the performance of prototype-based \ac{xai} methods, especially for time series data. This absence hinders researchers and practitioners from determining the best prototype-based explainability method for specific applications. Without reliable benchmarks, evaluations tend to be subjective and inconsistent, leading to fragmented findings and slow progress in the field~\cite{Weber24}. 
Researchers frequently use human evaluations and user studies, respectively, for explanation methods~\cite{mohseni2020quantitative}. While this approach is a valuable complement to analysis~\cite{Hoffman2018}, relying on human judgment to identify correct explanations poses challenges. The concept of an {\textquotedblleft accurate\textquotedblright} explanation can vary significantly based on the task, and humans may not consistently assess the quality reliably~\cite{Wang.19}. Recent studies reveal that quantitative evaluations of explainable methods also have significant limitations. It is important to be cautious about the metrics used to evaluate explanation methods, as many existing metrics may not accurately reflect the user understanding or model behavior~\cite{hase-bansal-2020-evaluating}.
Existing frameworks and metrics provide insights for assessing various explanation types, but they often lack the specificity needed to evaluate prototype-based explanations, especially in time series analysis. For a review on existing methods, see Section~\ref{chap:relatedwork} and Appendix~\ref{chap:evaluationframeworks}. The limited benchmarks targeting time series data highlight a critical gap in the current landscape of \ac{xai} evaluations. Prototype-based methods are promising for their interpretability and relevance for time series data. However, the absence of tailored evaluation frameworks makes it difficult to ascertain their effectiveness and reliability in practice. 

Our contributions to the field of evaluating XAI methods are:
\begin{itemize}
    \item A standardized and robust framework for automated, fair, comprehensive and reproducible evaluation of prototype-based explainability methods.
    \item Introduction of two additional conceptual attributes for prototypes that were not covered by \citet{Nauta.2023}.
    \item Interpretation and adaptation of all conceptual attributes specifically for prototypes.
    \item Definition of corresponding metrics for all conceptual attributes for prototypes.
    \item Facilitation of comparisons across different data types, such as time series data and image data.
    \item Assistance in selecting explanations for user studies without replacing them, to minimize the overall financial and time costs associated with conducting large experiments.
\end{itemize}
% This work aims to create a standardized framework for fair, comprehensive, and reproducible evaluation of prototype-based explainability methods. It facilitates comparisons for different data types including time series and image data, while rendering the Co-12 properties of \citet{Nauta.2023} quantifiable for prototype-based models. The proposed framework assists in selecting explanations for user studies, without replacing them, to minimize the  financial and time costs associated with conducting such large experiments.

We first provide a comprehensive background on existing evaluation methods for \ac{xai}, especially prototype-based methods, highlighting current challenges. In Section~\ref{chap:methodology}, we introduce the criteria and associated metrics forming the foundation of our benchmark. We then present  a practical use case to illustrate the applicability of our proposed framework in Section~\ref{chap:Experiments} and demonstrate how different prototype methods can be evaluated with the established criteria to reveal the strengths and weaknesses of the prototype methods. Finally, we discuss the evaluation results and functionality of the benchmark tool in Section~\ref{chap:discussion} and close the paper by summarizing our findings and suggesting directions for future research in prototype-based \ac{xai} evaluations.

%\begin{itemize}
% \item Deep Learning for Case-Based Reasoning through Prototypes:
%A Neural Network that Explains Its Predictions
%\url{https://ojs.aaai.org/index.php/AAAI/article/view/11771}
% possibly the first publication on prototypes with autoencoder 
%\item This Looks Like That: Deep Learning for Interpretable Image Recognition
% \url{https://arxiv.org/abs/1806.10574}
%\end{itemize}

\section{Related Work} 
\label{chap:relatedwork}
%\subsection{Survey of \ac{xai} method Evaluations and Key Properties for Assessment}
Various reviews, qualitative and quantitative evaluations, and benchmarks exist for \ac{xai}. Some reviews aid in selecting the appropriate explanation type, such as feature importance, counterfactuals, or prototype-based explanations. There exists a first comprehensive literature review on \ac{xai} for time series, distinct from image classification~\cite{Theissler2022ExplainableAF}.
Other surveys focus on assessing \ac{xai} approaches. The \emph{Co-12 properties} ~\cite{Nauta.2023}, a set of 12 conceptual attributes such as Compactness and Correctness, provide a foundation for evaluating explanation quality. Therefore, Chapter~\ref{chap:methodology} provides more detailed information on the Co-12 properties, as they will be used to structure the proposed benchmark.
Evaluation methods based on these Co-12 properties for part-prototypes have been proposed~\cite{Nauta.2023b}. \emph{Part-prototypes} cover a limited number of input dimensions, e.g., only show part of an image, and not entire input samples, as done with \emph{non-part prototypes}.
However, these suggestions for the usage of existing evaluation methods are specifically designed for part-prototypes in computer vision and may not assess all 12 properties for other data types and for non-part-prototypes, for example time series prototypes.
Benchmarks for evaluating \ac{xai} methods on selected datasets also exist. Out of a couple of such studies, one evaluates \ac{xai} methods for time series data and introduces two novel approaches such as perturbation and sequence analysis~\cite{Schlegel19}. But it focuses only on Saliency Map methods, such as LRP, LIME, DeepLift, and SHAP, excluding prototype methods. These evaluation methods are employed to validate XAI explanations, including LIME, SHAP, and Path Integrated Gradients (PIG), specifically adapted for multivariate time series classification in the maritime domain~\cite{VEERAPPA2022101539}.

%\subsection{Frameworks for Evaluating and Comparing \ac{xai} methods}
Several frameworks have emerged to compare and evaluate \ac{xai} methods, with \emph{Quantus}~\cite{quantus} being the only one applicable to prototype-based \ac{xai}. A detailed description of further evaluation frameworks can be found in Appendix~\ref{chap:evaluationframeworks}. Quantus provides 37 evaluation metrics, covering a wide range of explanation types, including feature importance, heatmaps, localization, prototypes, and decision trees/rules methods. The metrics are categorized into faithfulness, robustness, localization, complexity, randomization, and axiomatic properties. However, some metrics represent different versions of the same concept, most are only applicable to attribution methods—not for prototypes—and the library lacks public benchmarks.
The most common automated approaches for the quantitative evaluation of \ac{xai} methods were investigated by \citet{Le2023BenchmarkingEA}, who identified and analyzed 17 toolkits regarding the coverage of the Co-12 properties of \citet{Nauta.2023} and whether they yield consistent results. While multiple toolkits exist for standard explanation tasks such as feature importance and heatmaps, benchmarks for assessing prototype methods are rarely found. The analysis noted that time series data is supported only by Quantus~\cite{quantus}, which offers limited assessment for prototype explanations, focusing on image data with just three metrics. Despite a year and a half having passed since the publication of \citet{Le2023BenchmarkingEA}, no new benchmarks for prototype methods have been proposed.

While significant progress in evaluating and benchmarking \ac{xai} methods, challenges remain, particularly for prototype-based approaches in time series data. To address these issues, we develop a dedicated benchmark framework \emph{ProtoScore}, to evaluate prototype-based \ac{xai} methods across various data types, but especially for time series data. Our framework integrates the Co-12 properties to ensure a comprehensive assessment of explanation quality. The use of established properties allows for meaningful comparison of prototype methods to other \ac{xai} approaches evaluated along the same dimensions.
    
\section{Methodology}
\label{chap:methodology}
Our benchmark, \emph{ProtoScore}, is designed to evaluate trained prototype models, focusing on their explainability properties. To define specific metrics and assess the quality of prototypes adequately, certain assumptions are necessary. These assumptions include that a prototype closely resembles or is identical to a data instance, and that the set of prototypes represents the entire dataset~\cite{Molnar.2022}. Consequently, prototype methods are excluded, which optimize prototypes for proximity to the decision boundary, as discussed by \citet{wang2023learningsupporttrivialprototypes}, as they do not represent the dataset. The focus of the benchmark is strictly on the technical aspects of prototypes, using the latent space of the prototype model. Therefore, only models that incorporate a latent space can be evaluated with our benchmark tool. To begin, we define essential quantities that are necessary to evaluate the prototypes in Section~\ref{sec:definitions}. The development of ProtoScore is primarily based on the properties outlined by \cite{Nauta.2023, Nauta.2023b} for comprehensive and comparable evaluation. These properties are described in Section~\ref{sec:criteria}, accompanied by two additional properties we deem necessary for prototype methods. We provide metrics evaluating each property, which are specially adapted for prototypes, yielding individual scores that are then averaged to produce a total score. The implementation details are described in Section~\ref{sec:implementation}.

\subsection{Preliminaries}
\label{sec:definitions}
We begin by defining key concepts that will be used to establish metric later. Given a set of input samples
%data%
\begin{equation*}
\mathcal{D} =\{ x_1, x_2, \dots, x_N\} \quad \text{with } x_i \in \mathbb{R}^d \text{ and } |\mathcal{D}|=N~,
\end{equation*}
we define an encoder function that maps each data point $ x_i$ to its latent representation $z_i$ in a lower-dimensional space 
\begin{equation}
f: \mathbb{R}^d \to \mathbb{R}^n  \quad x_i \mapsto z_i \quad \text{ with } d \gg n~,
\label{eq:encoder}
\end{equation}
%latent representation%
where $d$ is the input space dimension and $n$ is the latent space dimension. The latent representation of $\mathcal{D}$ is given by
\begin{align*}
    \mathcal{D}_{\mathrm{enc}} &= \{f(x_1), f(x_2), \dots, f(x_N)\}=\{z_1, z_2,\dots, z_N\}   \\ &\text{with } x_i \in D \text{ and } z_i \in \mathcal{D}_{\text{enc}}~.
\end{align*}
In this latent space we identify the set of prototypes, where in some prototype methods $\mathcal{P}\subseteq \mathcal{D}_\mathrm{enc}$, by
\begin{equation*}
\mathcal{P} = \{ p_1, p_2, \dots, p_M \}~, \quad \text{with } p_i \in \mathbb{R}^n \text{ and } |\mathcal{P}|=M~.
\end{equation*}
The prototype classifier network for predicting $y_i$ from $z_i$ is defined as
\begin{equation}
h: \mathbb{R}^{n} \to \mathbb{R}\quad z_i \mapsto y_i~,
 \label{eq:prediction}
\end{equation}
resulting in the overall output given by $y_i=(h\circ f)(x_i)$.
The decoder function is used to transform each data point or associated prototype from the latent representation back to the input space
\begin{equation}
g: \mathbb{R}^n \to \mathbb{R}^d  \quad z_i \mapsto x_i~, \quad p_j \mapsto x_{p_j}~.
\label{eq:decoder}
\end{equation}
In the latent space, the dataset $\mathcal{D}_\mathrm{enc}$ can be grouped into clusters $C_i$, resulting in the set of clusters
\begin{equation*}
       \mathcal{C} = \{C_1,C_2, \dots, C_K\}\quad \text{with }  \bigcup_i C_i = D_\mathrm{enc} \quad \text{and } |\mathcal{C}|=K~. 
\end{equation*}
The average distance from a point $z_i$ to a cluster $C_j$ is defined as
\begin{equation*}
    a(z_i,C_j) = \frac{1}{|C_j\setminus\{z_i\}|} \sum_{\substack{z \in C_j\setminus\{z_i\}}} \Vert z_i-z\Vert_2~, %\quad \text{with } z_i\in C_j~,
\end{equation*}
where $\Vert z_i-z\Vert_2$ indicates the Euclidean distance. In this calculation $z_i$ is excluded to avoid considering the distance from a data point to itself. The motivation to use the Euclidean distance instead of other metrics like the ones suggested in \citet{hanawa2021evaluation} is its straightforward interpretation. This is particularly relevant in a clustering context, where the aim is to minimize the geometric distance between data points. The Euclidean distance is robust and less sensitive to outliers compared to other distance metrics. Moreover, metrics like cosine similarity focus only on the direction and ignore the magnitude of the distance between points. Thus, two points can be far apart in space yet share the same direction.
%For the distance $d(z_i, z) $ the Euclidean distance is employed, defined as
%\begin{equation}
%d(\mathbf{z}_i, \mathbf{z}_j) = \sqrt{\sum_{l=1}^{n} (z_{i,l} - z_{j,l})^2} = \Vert z_i-z_j\Vert_2~.
%\label{eq:euclidean}
%\end{equation}
The Silhouette score \cite{ROUSSEEUW198753} is calculated using the mean intra-cluster distance $a(z_i, C_k)$, where $z_i\in C_k$, and the mean nearest-cluster distance $a(z_i, C_{m})$, where 
\begin{equation*}
C_{m}\in \underset{l \neq k}{\text{argmin}} \{a(z_i, C_l)\}
\end{equation*}
indicates the nearest Cluster to $z_i$. The Silhouette score is then defined as
\begin{equation}
s(z_i) = \frac{a(z_i, C_{m}) - a(z_i, C_k)}{\max \{a(z_i, C_m), a(z_i, C_k)\} }~.
\label{eq:silhouettescore}
\end{equation}
The Silhouette score is advantageous compared to other metrics as it quantifies the separation of clusters by assessing both cohesion (how close instances within the same cluster are) and separation (how far apart instances in different clusters are). Many other metrics focus exclusively on one aspect, resulting in an incomplete assessment of clustering quality. The Silhouette score is similar to the MMD-Critic Score~\cite{NIPS2016_5680522b} when using a kernel function such as the Euclidean distance in the MMD-Critic. Both assess the two mentioned aspects, though they have different scaling effects.
The clustering of data points in the latent space is optimized by minimizing the average Silhouette score over all data points $z_i$ using k-means clustering. This clustering will be used as baseline for assessing the prototypes. 
\begin{comment}
The clustering of the data points in the latent space is optimized by minimizing the average \emph{Silhouette score} \cite{ROUSSEEUW198753} over all data points $z_i$ using k-means clustering. This clustering will be used as baseline for assessing the prototypes. The Silhouette score for a single point is defined as
\begin{equation}
s(z_i) = \frac{b(z_i) - a(z_i)}{\max{(a(z_i), b(z_i))}}~,
\label{eq:silhouettescore}
\end{equation}
where $a(z_i)$  is the mean intra-cluster distance, %erklärung -1 
\begin{equation*}
a(z_i) = \frac{1}{|C_j|-1} \sum_{\substack{z \in C_j \ z \neq z_i}} d(z_i, z)\quad \text{with } z_i\in C_j~,
\end{equation*}
and describes how well the data point $z_i$ matches to its assigned cluster $C$ by calculating the average distance from this point to all other points in the same cluster.
Additionally, $b(z_i)$ is the mean nearest-cluster distance 
\begin{equation*}
b(z_i) = \min_{ l \neq j} \left( \frac{1}{|C_l|} \sum_{z \in C_l} d(z_i, z) \right)
\end{equation*}
that describes how well the data point $z_i$ matches the closest cluster $C_l$ with $l \neq j$.
\end{comment}
The centroid of cluster $C_i$ is calculated via
\begin{equation}
\mu_{i} = \frac{1}{|C_{i}|} \sum_{x \in C_{i}} x~, \quad \text{where } C_i \in \mathcal{C}~.
\label{eq:centroid}
\end{equation}
The set of centroids can then be expressed as
\begin{equation*}
    \mathcal{M} = \{\mu_1,\mu_2, \dots, \mu_K\} \quad \text{with } |\mathcal{M}|=|\mathcal{C}|=K~.
\end{equation*}
These clusters and centroids are used to assess the prototype quality as described in the following.

\subsection{Metrics}
\label{sec:criteria}
The Co-12 properties by \citet{Nauta.2023} are categorized into three main dimensions: Content, Presentation, and User. However, the properties lack practical usage for accessing prototype explanations. Therefore, our framework extends these conceptual attributes by providing applicable metrics to assess prototype quality as described in the following subsections. In our ProtoScore benchmark, a constrained set of those properties is used, as listed in Table~\ref{tab:criteria}. Since our benchmark focuses on testing the technical properties of prototypes, we excluded criteria that are not assessed automatically and are better suited for user studies. In total, seven of the Co-12 properties are evaluated. Two of the five properties not considered are inherently satisfied by design or identical across the prototype methods (Completeness, Composition), and one is not applicable (Controllability) for prototype models, eliminating the need for their evaluation. The remaining two properties (Context, Coherence) are not automatically assessable and thus, user studies are needed. This strategic exclusion of certain properties is grounded in their specific nature and context, thus the excluded properties and their exclusion are described in further detail in the following.

Completeness measures how well the explanation captures the model's decision-making process. It consists of Rea\-son\-ing-\-Com\-plete\-ness and Output-Completeness. Reasoning-Completeness indicates the degree to which the explanation describes the entire internal workings of the model. Reasoning-Completeness is inherently met by ante-hoc prototype models, as they provide an explanation that is equivalent to the model itself.
Output-Completeness focuses on how much of the output is explained within the explanation, ensuring that it sufficiently explains the prediction made by the model. Output-Completeness is also inherently satisfied by design in non-part prototype models.
Composition evaluates the presentation format and organization of explanations, focusing on how information is conveyed rather than the content itself. Since we only evaluate one explanation type---prototypes---and the design choices for displaying prototypes are independent of the prototype method, this aspect is not included in the benchmark. Controllability refers to users' ability to interact with and modify explanations. Since current prototype models are typically static and user-independent, no meaningful evaluation of this property is possible.
Context considers the user's needs, expertise, and specific application, emphasizing the importance of tailoring explanations to different users. This property cannot be evaluated technically. Instead, user studies can provide qualitative feedback on the relevance and effectiveness of explanations in real-world scenarios.
Coherence assesses the alignment of explanations with users' prior knowledge and beliefs, ensuring plausibility and reasonableness. This property cannot be evaluated automatically, but visualization of prototypes can help determine their alignment with common understanding.

We define two additional criteria to be incorporated into the benchmark, since the Co-12 insufficiently capture all aspects of prototypes, particularly concerning the latent space quality and representativeness of the prototypes, which are properties specific to prototype methods.
Each criterion is evaluated using metrics that are adopted for prototypes. All metrics provide values from $0$ to $1$, where $0$ indicates poor prototype quality and $1$ represents the highest quality. The evaluated criteria, as listed in Table~\ref{tab:criteria}, with our defined metrics, will be presented in the following.
\begin{table*}[t]
    \centering
    \caption{Criteria and metrics used in the benchmark tool.}
    \begin{tabular}{@{}ccll@{}}
        \toprule
        \textbf{Co-12 Property} & & \textbf{Criterion} & \textbf{New Defined Metric} \\ \midrule
        \multirow{8}{*}{Co-12} &  \multirow{5}{*}{  Content}  &   \textbf{Correctness} &   Fidelity\\ 
        &   &   \textbf{Consistency} &   Difference of prototypes between runs \\ 
        &  &   \textbf{Continuity} &   Stability of explanations under input perturbations  \\ 
        &  &   \textbf{Contrastivity} &   Distance of prototypes  \\ 
        &  &   \textbf{Covariate Complexity} &  Silhouette score of prototypes \\ %\cline{2-4}
        &\cellcolor{gray!20}& \textbf{Compactness} \cellcolor{gray!20}& Number of prototypes \cellcolor{gray!20}\\ 
        & \multirow{-2}{*}{\cellcolor{gray!20} Presentation} & \textbf{Confidence} \cellcolor{gray!20}& Distance of sample to corresponding prototype \cellcolor{gray!20} \\ %\cline{2-4}
        \midrule
        \multirow{2}{*}{New} & & \textbf{Input Completeness} & Ratio of representative prototypes to centroids  \\ 
        & & \textbf{Cohesion of Latent Space} & Silhouette score of clusters   \\ \bottomrule
    \end{tabular}
    \label{tab:criteria}
\end{table*}

\subsubsection{Correctness}
Correctness measures how accurately explanations reflect the model's reasoning and predictions. In prototype models, Correctness is linked to the model's ability to explain decisions through visualizations of learned prototypes.
Correctness is often evaluated using synthetic datasets where the true prototypes are known in advance. Measures for the Correctness of prototypes then include the number of found prototypes and the distance between found and true prototypes. However, since this type of evaluation compares the explanation and the dataset, this evaluation can yield low Correctness scores if the model is not well-adapted to the data, despite the explanations aligning with the model's reasoning, i.e., being correct.
An alternative evaluation method for Correctness, which is used here, measures the agreement between the output of the predictive model for an input sample and the output of the predictive model for the corresponding prototype, also known as \emph{fidelity}. For ante-hoc prototype methods, the fidelity only depends on the autoencoder's reconstruction loss, otherwise the extracted prototype reflects exactly the models behavior.
Thus, to rate the Correctness for the set of clusters $C$ in the latent space, we extract all data points $x_i$ from each cluster $C_j$ along with the corresponding prototypes $p_j$, and transform them into the input space using the given decoder defined by Equation~\eqref{eq:decoder}. The prediction for the data points and input space prototypes can then be computed using the encoder and classifier defined in Equation~\eqref{eq:encoder} and Equation~\eqref{eq:prediction}. Correctness is defined as the ratio of all matching predictions between the data points $ y_i $ and their respective prototype $ y_{p_i}$ to the total number of data points $N$,
\begin{equation*}
\text{CR} = \frac{1}{N} \sum_{i=1}^{N} \mathbb{I}_{\{y_{p_i}\}}(y_i)~,
\end{equation*}
where $\mathbb{I}_{\{y_{p_i}\}}(y_i)$ is the indicator function being one if the condition $y_{p_i} = y_i$ holds, otherwise it is zero.

\subsubsection{Consistency}
Consistency ensures that identical inputs produce the same explanation, promoting stability and reliability across instances and runs. It emphasizes implementation invariance, so that explanations should not vary with specific model details. Two models generating the same output for all inputs should provide the same explanations, regardless of their architectures. Since prototype models are typically deterministic, the same input should yield the same explanation. However, nondeterminism can arise from factors such as model initialization and random seeds. Evaluating latent distances of prototypes across different runs can indicate the stability of the model's explanations under varying conditions.
Consequently, we employ additional models trained under the same conditions as the model being tested. Prototypes from these models are decoded into the input space using Equation~\eqref{eq:decoder} with their respective decoder functions and  encoded into the latent space of the model under evaluation using Equation~\eqref{eq:encoder} to allow for comparison. The Euclidean distances between the prototypes of the new models $p_\mathrm{new}$ and the baseline model $p_\mathrm{base}$ are calculated and averaged over all prototypes $M$ and all new models $N_R$. The Consistency score is then given by the inverted normalized mean distance % as defined in Equation~\eqref{eq:euclidean}, which gives
\begin{equation*}
    \text{CS}= \exp{\left(-\frac{1}{M\cdot N_R}\sum_{i=1}^M\sum_{j=1}^{N_R}\Vert p_{\text{base}_{i}} - p_{\text{new}_{i,j}}\Vert_2\right)}~.
\end{equation*}
This score ranges from $0$ to $1$, with $p_{\text{new}_{i,j}}$ represented in the latent space of $p_{\mathrm{base}_i}$ and corresponding to the same cluster $C_i$.
%\begin{equation*}
%\Delta_\mathrm{runs} = \frac{1}{M\cdot N_M}\sum_{j=1}^{N_R}\sum_{i=1}^M\Vert p_{\text{base}_{i}} - p_{\text{new}_{i,j}}\Vert_2~.
%\end{equation*}
%The Consistency score is then given by the inverted normalized mean distance,
%\begin{equation*}
%    \text{CS}= \exp{(-\Delta_{\text{runs}})},
%\end{equation*}
%to be in the range of $0$ to $1$. Here, $p_{\text{new}}$ are in the \emph{same} latent space of $p_\mathrm{base}$.

%Consistency (Dasgupta et al., 2022 https://arxiv.org/abs/2202.00734): measures the probability that the inputs with the same explanation have the same prediction label 

\subsubsection{Continuity}
Continuity measures the stability and generalizability of the explanation function, emphasizing that similar inputs should produce similar explanations. This property focuses on the model's ability to maintain consistent explanations even with slight input perturbations. We evaluate Continuity by introducing small Gaussian noise to the input dataset $\mathcal{D}$, resulting in a noisy dataset
\begin{align*}
&\mathcal{D}_{\text{noised}} = { x_1^\prime, x_2^\prime, \dots, x_N^\prime } \\ &\text{where } x_i^\prime = x_i + \epsilon_i  \text{ with } x_i \in D, \epsilon_i \sim \mathcal{N}(0,\sigma^2 I)~.
\end{align*}
Here, $\sigma$ is set to $5~\%$ of the average range of sample values and $I$ is the identity matrix.
Both $\mathcal{D}$ and $\mathcal{D}_{\text{noised}}$ are encoded by the encoder function~\eqref{eq:encoder}.
For each data point $x_i$ or noisy data point $x_i\prime$ in the original dataset, we identify its closest prototype of the prototype set $\mathcal{P}$ by
\begin{equation}
p_{x_i} = \arg\min_{p_j \in \mathcal{P}} \Vert f(x_i) - p_j\Vert_2~.
\label{eq:closest_prototype}
\end{equation}
We assess the model’s prototype stability to input perturbation by computing the average distance between prototypes of the original and noised datasets, normalizing and inverting this distance to obtain the Continuity score
\begin{equation*}
    \text{CN}=\exp{\left(- \frac{1}{N} \sum_{i=1}^N \Vert p_{x_i} - p_{x_i^\prime} \Vert_2\right)}\quad\text{with } x\in \mathcal{D},\ x^\prime \in \mathcal{D}_\mathrm{noised}~.
\end{equation*}
%We assess the stability of the model’s prototype to input perturbation by computing the average distance
%\begin{equation*}
%\Delta_\mathrm{noise} = \frac{1}{N} \sum_{i=1}^N \Vert p_{x_i} - p_{x_i^\prime} \Vert_2 \quad \text{with } x\in \mathcal{D},\ x^\prime \in \mathcal{D}_\mathrm{noised}~.
%\end{equation*}
%Again we normalize and invert the distance to receive the Continuity score
%\begin{equation*}
 %   \text{CN}=\exp{(-\Delta_{noise})}~.
%\end{equation*}
%Continuity (Montavon et al., 2018 https://arxiv.org/pdf/1706.07979.pdf): captures the strongest variation in explanation of an input and its perturbed version 

%Relative Input Stability (RIS) (Agarwal, et. al., 2022 https://arxiv.org/pdf/2203.06877.pdf): measures the relative distance between explanations ex and ex' with respect to the distance between the two inputs x and x' 

\subsubsection{Contrastivity}
Contrastivity measures an explanation's ability to distinguish between different classes or targets.
In ante-hoc prototype models, this feature is integrated into the design, where different classifications reflect distinct reasoning processes and corresponding explanations/prototypes. Thus, the distance between different prototype representations indicates their discrimination strength. We calculate Contrastivity as the mean distances between prototypes in the latent space using the Euclidean distance according to
\begin{equation*}
    \text{CT}=\frac{1}{M(M-1)}\sum_{p_i,p_j \in P, i\neq j}\Vert p_i-p_j\Vert_2~.
\end{equation*}

\subsubsection{Covariate Complexity}
Covariate Complexity refers to the difficulty users may have in understanding an explanation, i.e., the shown prototypes. The complexity or understandability of these visual features can be quantified through user studies that evaluate human interpretation. The evaluation of Covariate Complexity often involves measuring the {\textquotedblleft{purity}\textquotedblright} of the clusters with association to the closest prototype. A high purity score indicates that the explanations highly resemble a particular concept. Thus, we can assess this property automatically by calculating the Silhouette score as defined in Equation~\eqref{eq:silhouettescore}, here calculated for the prototypes included as elements to their respective cluster set. The Silhouette score rates if the prototypes are close to the data points that encode a similar concept, but are far away from data points with different concepts. While automated purity metrics are valuable, they should be complemented by user studies to ensure that the prototypes accurately reflect human understanding.

\subsubsection{Compactness}
Compactness refers to the size of the explanation provided by a model, emphasizing that explanations should not overwhelm the user. It is evaluated by counting the number of prototypes.
The compactness is calculated by taking the total number of prototypes $M$ and normalizing it as 
\begin{equation*}
\text{CP} = \exp{\left((-M + 1)\cdot a_\mathrm{normalize}\right)}~,
\end{equation*}
where $a_\mathrm{normalize}=0.08$ is a normalization hyperparameter that ensures the $0-1$ range for typical numbers of prototypes. The factor is chosen to balance the score, resulting in a mediocre score of 0.5 for ten prototypes, with fewer prototypes yielding better scores and more prototypes leading to lower scores.

\subsubsection{Confidence}
Confidence refers to the degree of certainty a model has regarding its predictions and explanations. It encompasses the availability of probability information related to the model's output and the reliability of those probability estimates. It is typically expressed as a probability score reflecting how certain the model is about a classification. Confidence scores derived from models trained with softmax and cross-entropy loss can be assessed for their reliability. We distinguish between different types of confidence: classification confidence (related to the predicted class) and explanation confidence (related to the reliability of the explanation generation). %, and out-of-distribution detection confidence. 
For prototype methods---given that the classification is generated via the class of the nearest prototype---classification confidence and explanation confidence are the same.
Confidence is measured by calculating the average distance of the data points in the latent space $z_i$ to the closest prototype $p(x_i)$, where the closest prototype $p(x)$ can be found by means of Equation~\eqref{eq:closest_prototype}. 
Then the Confidence score is given by
\begin{equation*}
    \text{CF}=\exp{(-\frac{1}{N} \sum_{i=1}^N \Vert z_i - p_{x_i}\Vert_2)}~.
\end{equation*}

\subsubsection{Input Completeness}
Input Completeness is an extension of Reasoning Completeness and Output Completeness, measuring how well prototypes represent the training data. It assesses the proportion of training data points adequately captured by the prototypes, indicating the model's ability to generalize to unseen instances. A high Input Completeness value suggests that prototypes are well-distributed and align closely with the data distribution, while a low value may indicate underrepresented regions, leading to poor generalization.
Input Completeness is assessed by the ratio of representative prototypes to the number of clusters. Therefore, a prototype $p_{i}$ is considered representative of  $C_{j}$  if its distance 
to the cluster’s centroid $\mu_j$ is smaller than the average intra-cluster distance $a(\mu_{j}, C_{j})$. We define the set of clusters that have at least one representative prototype as
\begin{equation} 
\label{eq:representative-prototype}
C_{\text{rep}} = C_{j} \in C \mid \exists p_{i} \in \mathcal{P} \text{, such that } \Vert p_{i} - \mu_{j} \Vert_{2} < a(\mu_{j},C_{j})~,
\end{equation} 
where the average intra-cluster distance is calculated by the average Euclidean distance from every point in  $C_{j}$  to its centroid $\mu_{j}$ according to
\begin{equation*}
a(\mu_{j}, C_{j}) = \frac{1}{|C_{j}|} \sum_{z \in C_{j}} \Vert \mu_{j} - z \Vert_2~.
\end{equation*}
Then, the Input Completeness score is given by
\begin{equation*}
\text{IC} = \frac{|C_{\text{rep}}|}{|C|}~,
\end{equation*}
which is the relative frequency of clusters that have a representative prototype, i.e., these clusters satisfy the acceptance criterion \eqref{eq:representative-prototype} and thus, comprise at least one prototype.

\subsubsection{Cohesion of Latent Space}
Not only the prototypes are evaluated but also the latent space is assessed because it is essential as the foundation for calculating the prototypes. The quality of the latent space is crucial as it directly impacts the representation of the data and the effectiveness of the prototypes derived from it. This criterion evaluates how well the latent space organizes the training data into distinct clusters. A well-structured latent space produces clusters that are easily separable from one another, allowing for clear differentiation between various categories or classes within the data. Effective clustering ensures that similar data points are grouped together, enhancing the model's ability to generalize and make accurate predictions on new data.

To quantify the clustering ability of the latent space, we use the Silhouette score defined in Equation~\eqref{eq:silhouettescore}. This metric evaluates how similar an object is to its own cluster compared to other clusters. A higher Silhouette score indicates better-defined clusters. The Cohesion of Latent Space (CLS) Score is defined as the average Silhouette score across all clusters $C_i$ with centroid $\mu_i$
\begin{equation*}
    \text{CLS} = \frac{1}{\lvert \mathcal{C} \rvert} \sum_{i=1}^n s(\mu_i) \quad \text{for } C_i \in \mathcal{C} \text{ and } \mu_i \in M~.
\end{equation*}

\subsection{Implementation of the Benchmark Framework}
\label{sec:implementation}

\begin{algorithm}[t]
    \caption{Benchmark Framework}\label{alg:benchmark}
    \begin{algorithmic}[1]
        \REQUIRE Encoder $f$, Decoder $g$, Prototype Classifier $h$, Dataset $\mathcal{D}=\{x_1,\cdots,x_N\}$, Labels, Prototypes $\mathcal{P}= \{ p_1, p_2, \dots, p_M \}$
        \ENSURE Prototype Quality Scores
        \STATE Transform input data into latent space: $z_i=f(x_i)$ (see Equation~\eqref{eq:encoder})
        \STATE Classify latent representations $z_i$ by means of ground-truth labels
        \STATE Cluster the data for each class using k-means by optimizing Silhouette score as defined in Equation~\eqref{eq:silhouettescore}
        \STATE Compute cluster centroids according to Equation~\eqref{eq:centroid} 
        \STATE Assign prototypes to each cluster resp. data point using Equation~\eqref{eq:closest_prototype}
        \STATE Calculate all prototype quality metrics listed in Table~\ref{tab:criteria}
        \STATE Compute a total score by averaging the individual scores with equal weighting
    \end{algorithmic}
\end{algorithm}
The benchmark framework (cf. Algorithm~\ref{alg:benchmark}) requires an encoder~\eqref{eq:encoder}, a decoder~\eqref{eq:decoder}, a trained classifier, a dataset with labels on which the autoencoder and prototype classifier network have been trained, and a list of learned prototypes. Prior to calculating the metrics for the specified criteria, several computations are required, as detailed in Section~\ref{sec:definitions}. The input data is transformed into a representation within a latent space using the provided encoder. Subsequently, the data points are categorized according to their ground-truth labels. For each class, clustering is performed through minimizing the Silhouette score defined in Equation~\eqref{eq:silhouettescore} and described in Section~\ref{sec:definitions} to identify the most appropriate number of clusters and the clusters themselves, respectively, using k-means. K-means is tested for $2-15$ clusters with the best performing number used, based on the Silhouette score. For each cluster, a centroid is computed as shown in Equation~\eqref{eq:centroid}, and prototypes are mapped according to Equation~\eqref{eq:closest_prototype}. %The Euclidean distance between each prototype and each centroid is then calculated to form a distance matrix. This distance matrix is used to find the nearest prototypes. to each cluster.% establish a mapping function from centroids to prototypes through the application of the Hungarian method~\cite{Kuhn1955TheHM}, which involves finding the optimal way to assign the set of prototypes $P$ to the set of clusters $C$ to minimize total distance between the prototypes and cluster centroids. % and therefore effectively address the linear sum assignment problem, a fundamental combinatorial optimization problem~\cite{Burkard12}. This process results in an mapping of prototypes to clusters. 
Finally, the metrics outlined in Section~\ref{sec:criteria} are computed to evaluate each property individually, culminating in the calculation of an overall score that assesses the quality of the explanation. The complete implementation is available on GitHub\footnote{\url{https://github.com/HelenaM23/ProtoScore}}.

\section{Experiments}
\label{chap:Experiments}
In this section we present an exemplary use case to demonstrate the usage of ProtoScore and additionally a more detailed investigation of prototype methods on multiple datasets. The description of the used datasets and prototype methods is included in Appendix~\ref{chap:datasetsmethods}.

%\begin{table}[ht]
 %   \centering
 %   \caption{Results of the experiments: (CR) Correctness, (CS) Consistency, (CN) Continuity, Contrastivity (CT), Covariate Complexity (CC)  Compactness (CP), Confidence (CF), Input Completeness (ICM), Cohesion of Latent Space (LS)}
 %   \begin{tabular}{cc|cccccccccc|c}
 %       \toprule
 %       \textbf{dataset} & \textbf{Method} & \textbf{Val Loss} &   \textbf{Cr}  &   \textbf{CS} &   \textbf{CN}  &   \textbf{CT} &   \textbf{CC}  & \textbf{CP} & \textbf{CF} & \textbf{CV}  & \textbf{LS} &\textbf{Total} \\
 %       \midrule
 %       Wafer & MAP & MSE: $0.2$ & &&&&&&&&&\\
 %       Wafer & MSP & MSE:$0.2$ & &&&&&&&&&\\
%        \\ \bottomrule
%    \end{tabular}
%    \label{tab:results}
%\end{table}

\subsection{Exemplary Use Case}
\begin{table*}[t]
    \centering
     \caption{Benchmark results for different models using the time series prototype methods MAP and MSP with the ECG200 dataset. The abbreviations in the column headings are the tested properties:
      (CR) Correctness, (CS) Consistency, (CN) Continuity, (CT) Contrastivity, 
      (CC) Covariate Complexity, (CP) Compactness, (CF) Confidence, (IC) Input Completeness, 
      (CLS) Cohesion of Latent Space.}
    \label{tab:results}
    \begin{tabular}{c|cc|c|ccccccccc|c}
        \toprule
        \textbf{Index} & \textbf{Dataset} & \textbf{Method} & \textbf{Val Loss}
        & \textbf{CR} & \textbf{CS} & \textbf{CN}
        & \textbf{CT} & \textbf{CC} & \textbf{CP}
        & \textbf{CF} & \textbf{IC} & \textbf{CLS}
        & \textbf{Total}\\
        \midrule
1 & ECG200 & MAP & MSE: $1.45$
            & 0.69 & 0.28
            & 0.98 & 0.43
            & 0.68 & 0.79
            & 0.67 & 0.67
            & 0.63 & 0.65 \\

2 & ECG200 & MAP & MSE: $0.70$
            & 0.78 & 0.34
            & 0.98 & 0.37
            & 0.66 & 0.79
            & 0.68 & 0.80
            & 0.64 & 0.67 \\
            
%2 & ECG200 & MAP & MSE: $0.60$
%            & 0.85 & 0.35
%            & 0.98 & 0.37
%            & 0.67 & 0.79
%            & 0.63 & 0.67
%            & 0.64 & 0.66 \\

3 & ECG200 & MAP & MSE: $0.55$
            & 0.85 & 0.40
            & 1.00 & 0.41
            & 0.65 & 0.79
            & 0.69 & 0.67
            & 0.66 & 0.68 \\
            
4 & ECG200 & MAP & MSE: $0.38$
            & 0.88 & 0.46
            & 1.00 & 0.30
            & 0.69 & 0.79
            & 0.61 & 0.57
            & 0.66 & 0.66 \\

5 & ECG200 & MAP & MSE: $0.26$
            & 0.88 & 0.46
            & 0.97 & 0.30
            & 0.69 & 0.79
            & 0.60 & 0.57
            & 0.64 & 0.66 \\ 
            
\cellcolor{gray!20}1 & \cellcolor{gray!20}ECG200 &\cellcolor{gray!20}MSP & \cellcolor{gray!20}MSE: $1.04$
            & \cellcolor{gray!20}1.00 & \cellcolor{gray!20}0.37
            &\cellcolor{gray!20}1.00 &\cellcolor{gray!20}0.49
            &\cellcolor{gray!20}0.50 &\cellcolor{gray!20}0.79
            &\cellcolor{gray!20}0.55 &\cellcolor{gray!20}0.00
            &\cellcolor{gray!20}0.60 &\cellcolor{gray!20}0.59\\

\cellcolor{gray!20}2 & \cellcolor{gray!20}ECG200 & \cellcolor{gray!20}MSP & \cellcolor{gray!20}MSE: $0.78$
            & \cellcolor{gray!20}1.00 & \cellcolor{gray!20}0.45
            & \cellcolor{gray!20}1.00 & \cellcolor{gray!20}0.46
            & \cellcolor{gray!20}0.50 & \cellcolor{gray!20}0.79
            & \cellcolor{gray!20}0.43 & \cellcolor{gray!20}0.00
            & \cellcolor{gray!20}0.52 & \cellcolor{gray!20}0.57 \\

\cellcolor{gray!20}3 & \cellcolor{gray!20}ECG200 & \cellcolor{gray!20}MSP & \cellcolor{gray!20}MSE: $0.36$
            & \cellcolor{gray!20}1.00 & \cellcolor{gray!20}0.55
            & \cellcolor{gray!20}0.95 & \cellcolor{gray!20}0.37
            & \cellcolor{gray!20}0.53 & \cellcolor{gray!20}0.79
            & \cellcolor{gray!20}0.47 & \cellcolor{gray!20}0.00
            & \cellcolor{gray!20}0.39 & \cellcolor{gray!20}0.56 \\

%\cellcolor{gray!20}4 & \cellcolor{gray!20}ECG200 & \cellcolor{gray!20}MSP & \cellcolor{gray!20}MSE: $0.30$
%            & \cellcolor{gray!20}0.34 & \cellcolor{gray!20}0.60
%            & \cellcolor{gray!20}0.90 & \cellcolor{gray!20}0.43
%            & \cellcolor{gray!20}0.57 & \cellcolor{gray!20}0.79
 %           & \cellcolor{gray!20}0.62 & \cellcolor{gray!20}0.80
 %           & \cellcolor{gray!20}0.64 & \cellcolor{gray!20}0.63 \\
\cellcolor{gray!20}4 & \cellcolor{gray!20}ECG200 & \cellcolor{gray!20}MSP & \cellcolor{gray!20}MSE: $0.30$
            & \cellcolor{gray!20}0.86 & \cellcolor{gray!20}0.60
            & \cellcolor{gray!20}0.86 & \cellcolor{gray!20}0.33
            & \cellcolor{gray!20}0.74 & \cellcolor{gray!20}0.79
            & \cellcolor{gray!20}0.66 & \cellcolor{gray!20}0.20
            & \cellcolor{gray!20}0.73 & \cellcolor{gray!20}0.64 \\
            
\cellcolor{gray!20}5 & \cellcolor{gray!20}ECG200 & \cellcolor{gray!20}MSP & \cellcolor{gray!20}MSE: $0.28$
            & \cellcolor{gray!20}0.96 & \cellcolor{gray!20}0.60
            & \cellcolor{gray!20}0.83 & \cellcolor{gray!20}0.38
            & \cellcolor{gray!20}0.61 & \cellcolor{gray!20}0.79
            & \cellcolor{gray!20}0.57 & \cellcolor{gray!20}0.00
            & \cellcolor{gray!20}0.73 & \cellcolor{gray!20}0.61\\
        \bottomrule
    \end{tabular}
\end{table*}
Users can rate the quality of the explanation of their prototype models using ProtoScore.
For this, they can either use their own dataset or one already provided in ProtoScore.
They can choose different prototype methods, train models with the selected dataset and run the benchmark. To use ProtoScore effectively, users should benchmark multiple models to obtain a diverse range of results. 
\begin{figure}[t]
    \centering
    \includegraphics[trim=0.8cm 22cm 0.5cm 0.8cm, clip,width=1\linewidth]{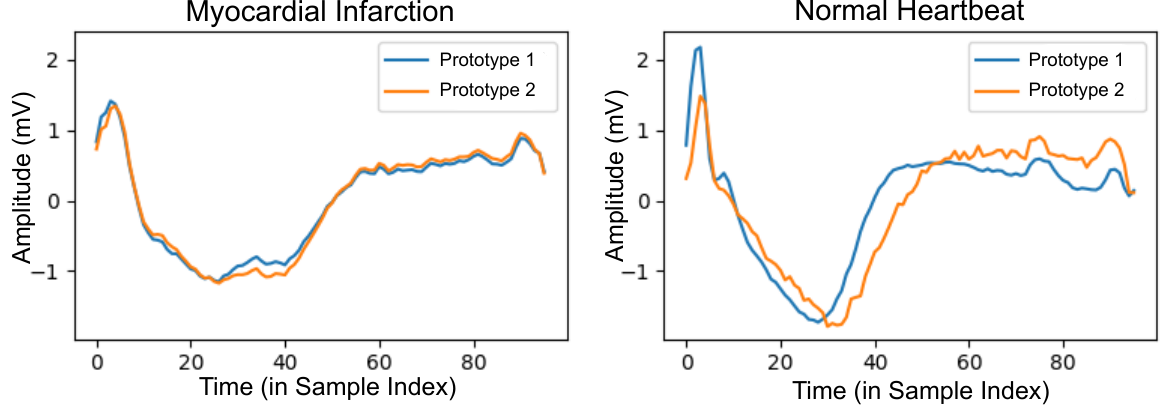}
    \caption{Exemplary prototypes for the MSP model with index 5. On the left, the prototype representing the abnormal class is displayed, while on the right, the prototype for the normal class, where specific conditions were detected, is shown.}
    \label{fig:prototypes}
\end{figure}
As an example, the results for two different time series prototype methods \emph{MAP} (Model-Agnostic Prototype)~\cite{pmlr-v189-obermair23a} and \emph{MSP} (Model-Specific Prototype)~\cite{gee2019explaining} are shown in Table~\ref{tab:results}. These methods, detailed in Appendix~\ref{chap:appendixmethods}, were trained on the ECG200 dataset~\cite{ECG200}, which is described in Section~\ref{chap:datasetsmethods} along with other datasets that are used in the following sections. An exemplary illustration of the resulting prototypes for the MSP model with index 5 is given in Figure~\ref{fig:prototypes}. Multiple models are trained for each method, as indicated by the index in the table, and evaluated against the prototype quality properties outlined in Section~\ref{chap:methodology}. The hyperparameters can be found in Appendix~\ref{chap:appendixhyperparameter}. The total score averages the individual scores with equal weighting to summarize the prototype quality. Alongside, the mean squared error (MSE) validation loss for each model is given, which provides context for the model performance. The methods are sorted by descending validation loss. Note that a score of $0$ indicates poor quality of the assessed prototypes, and a score of $1$ represents the highest quality possible. The Compactness (CP) is consistent across all models, as four prototypes are always calculated. This number can be adjusted by the user. It is set to four in this case based on experience suggesting that this quantity strikes an optimal balance between compactness and representativeness of the prototypes for the dataset.
% \begin{figure}[t]
%     %\centering
%      \begin{minipage}{0.42\textwidth}
%      \hfill 
%         \includegraphics[trim=0cm -2cm 3.8cm 5.9cm, clip, width=0.7\textwidth]{clustering_legend_map_ecg200.pdf}
%         \vspace{-3.5cm}
%     \end{minipage}
%     \begin{minipage}{0.5\textwidth}
%         \centering
%         \includegraphics[trim=0cm 0.5cm 0cm 0.2cm, clip, width=0.9\textwidth]{ECG200_3_epoch_low_lr_MAP.pdf}
%        % \subcaption{MAP method}
%         \label{fig:maplatentspace}
%     \end{minipage}
%      %
%     %\begin{minipage}{0.49\textwidth}
%      %   \centering
%       %  \includegraphics[width=\textwidth]{clustering_plot_msp_ecg200_30_epoch.pdf}
%      %   \subcaption{MSP method}
%       %  \label{fig:msplatentspace}
%    % \end{minipage}
%     \caption{Dimension-reduced representation of the latent space for the MAP model with index 1. Note that the dimensionality reduction distorts distances, causing centroids and prototypes to appear outside their respective clusters. Small dots represent input data, with lines connecting each prototype (circle) to its corresponding cluster centroid (square). Points are color-coded by class.}
%     \label{fig:latentspace}
% \end{figure}
\begin{figure}[t]
    \centering
    \includegraphics[trim=0.1cm 16cm 0.1cm 0cm, clip,width=1\linewidth]{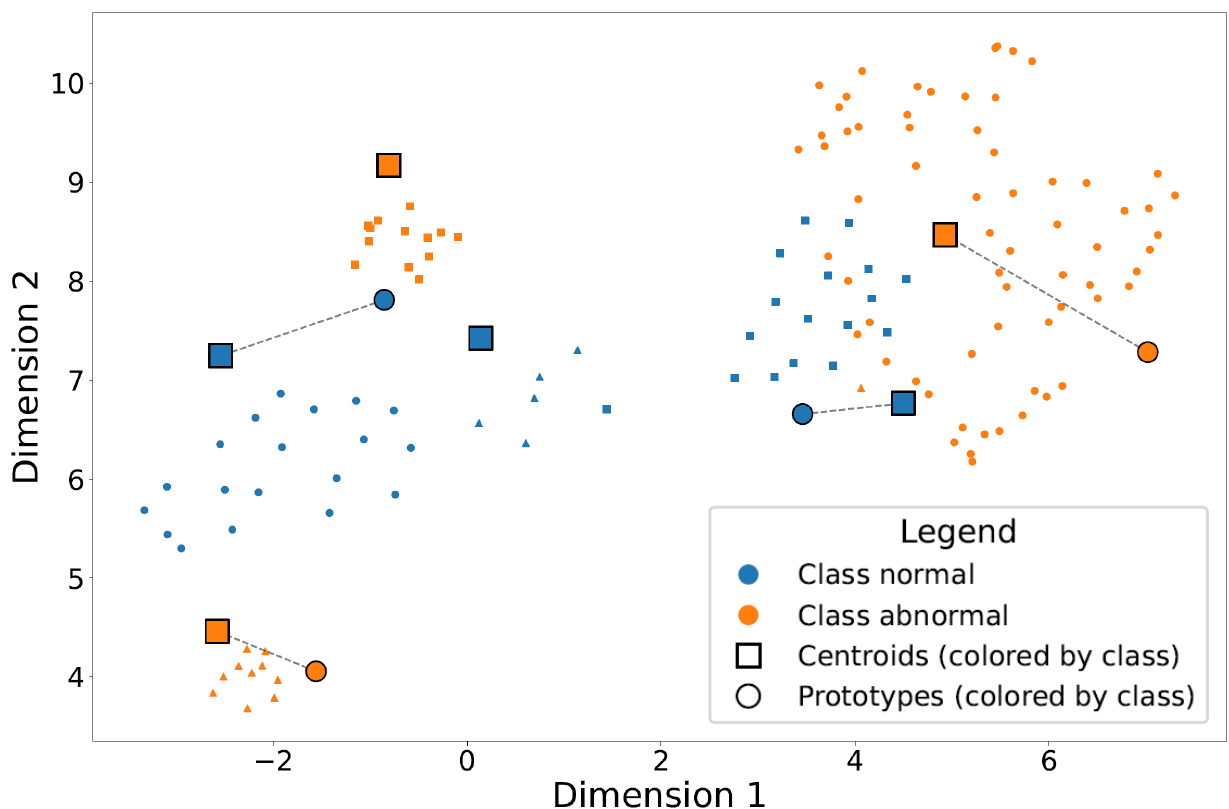}
    \caption{Dimension-reduced representation of the latent space for the MAP model with index 1. Note that the dimensionality reduction distorts distances, causing centroids and prototypes to appear outside their respective clusters. Small dots represent input data, with lines connecting each prototype (circle) to its corresponding cluster centroid (square). Points are color-coded by class.}
     \label{fig:latentspace}
\end{figure}

Figure~\ref{fig:latentspace} illustrates an exemplary latent space of the MAP model with index~1 (refer to Table~\ref{tab:results}). The latent space dimension is reduced using UMAP~\cite{umap}, a fast and scalable dimension reduction technique that preserves both local and global data structures for visualization, based on the assumption that the data lies on a locally connected Riemannian manifold. Note that this reduction distorts distances, rendering them incomparable in the plot. Thus, cluster centroids and prototypes may appear to be outside their respective clusters, even though they are actually part of them.
Analyzing the latent space reveals a strong correlation with the scores achieved by MAP's model 1 in Table~\ref{tab:results}. The overlap between the orange and blue clusters on the right side suggests that the challenges with Correctness (CR), Confidence (CF), and Cohesion of the Latent Space (CLS) may stem from this area. However, overall, the prototypes and clusters are well-separated, indicating strong performance in Contrastivity (CT), Covariate Complexity (CC), and Confidence (CF).
Also the representativeness of the prototypes can be seen. Six clusters are identified, with four featuring representative prototypes, as indicated by the connecting lines. This finding agrees with the Input Completeness score (IC) in the table.

Comparing the results in the table reveals that MAP achieves higher total scores across different validation losses compared to MSP. MAP's total score lies in the range of $0.65-0.68$, whereas MSP's  total score is is between $0.56$ and $0.64$. This indicates MAP's greater effectiveness in optimizing prototype quality, especially at lower validation loss values. Notably, MAP scores perform well in terms of properties such as Correctness (CR) ($0.69-0.88$) and Continuity (CN) ($0.98-1.00)$, which are essential to ensure that the generated prototypes explain the model's behavior correctly and are continuous. As MAP models improve in predictive accuracy (lower validation loss), Correctness (CR) and Consistency (CS) improve but Contrastivity (CT) decreases. This could be due to the also decreasing reconstruction loss but closer inspection reveals that the prototypes tend to be located closer together. The high total scores for MAP across all models suggest that it is a reliable method for generating high-quality prototypes. Even at higher validation loss values, MAP maintains a relatively stable total score, indicating consistent performance in prototype quality across various validation loss levels. The MSP method shows a more varied performance profile. While its total scores are generally lower than MAP's, MSP excels in specific properties for certain models. For example, MSP achieves perfect scores of $1.00$ in Correctness (CR) for model 1, model 2 and model 3. These high scores highlight MSP's ability to ensure that its prototypes align closely with the model's behavior. Moreover, MSP has two models with the highest Cohesion of Latent Space score (CLS) ($0.73$ for model 4 and model 5) and highest Consistency (CS) ($0.60$ for model 4 and model 5), indicating its ability to give a basis for calculating well separable prototypes and consistent prototypes over multiple runs. However, MSP exhibits weaknesses in other properties, such as Covariate Complexity (CC), Confidence (CF), Continuity (CN) and Input Completeness (IC), contributing to its overall lower total scores compared to MAP. The Input Completeness (IC) score of MSP is almost always $0.00$ for the ECG200 dataset because the distances from the prototypes to the respective cluster centroids are larger than the cluster spreads, making the prototypes unrepresentative for their respective clusters.

Overall, the results underscore the importance of considering multiple evaluation criteria when benchmarking prototype methods. While MAP outperforms MSP in the total score and is superior in most of the individual properties across different classification performances, MSP demonstrates strengths in certain properties that might be critical for specific applications, such as those prioritizing high Correctness and Consistency. 
Additionally, the inclusion of the validation loss in the evaluation provides an important context to interpret the results. A method with high prototype quality but high validation loss might indicate that its prototypes are well-constructed, but fail to generalize effectively in predictive tasks. Conversely, a method with low validation loss and good prototype quality represents a more balanced approach suitable for applications. Users aiming to apply ProtoScore effectively should carefully analyze both the total scores and individual property scores in relation to their specific objectives. For example, if model interpretability by less complex prototypes and high confidence about the allocation of prototypes to data points are priorities, MAP’s higher total scores and balanced property performance make it the preferable choice. Conversely, for tasks where Correctness and Consistency are paramount, MSP’s higher scores in CR and CS may offer advantages.

\subsection{Additional Experiments on More Datasets}
In addition to choosing between different prototype methods for a specific use case, ProtoScore can be used as a general benchmark for prototype-based \ac{xai} methods across different datasets. Corresponding results are listed in detail in Appendix~\ref{chap:appendixresults} in Table~\ref{tab:results2}. While our analysis focuses on univariate time series data, where $x_i$ represents a time series, ProtoScore is also applicable to multivariate datasets ($x_i$ represents all time series that belong to a data instance) and image data ($x_i$ represents an image), since our defined metrics apply across different data types.
Analyzing datasets like Wafer, SAWSINE, ECG200, FordA, FordB, and StarLightCurve highlights the applicability to both binary and multiclass classification. Evaluating MAP and MSP prototypes with our benchmark tool provides valuable insights. It reveals that MSP exhibits greater performance variability across different datasets, as evidenced in the ECG200 dataset. MSP achieves high property scores in specific instances but generally lower overall scores than MAP. MAP often achieves higher Correctness, Input Completeness and sometimes Continuity, while MSP excels in properties such as Cohesion of the Latent Space and sometimes in Confidence and Covariate Complexity. This emphasizes the importance of selecting methods based on the specific properties most relevant to the application.

Analyzing trends in loss values and their relationship to explanation quality reveals valuable insights. Generally, one might expect that as the validation loss decreases, the quality of explanations improves across the criteria. However, our results demonstrate that this relationship is not always consistent. For instance, while the validation loss for the MAP method on the ECG200 dataset in Table~\ref{tab:results} increases, certain properties such as Correctness and Consistency improve, while Contrastivity decreases. Moreover, datasets such as FordB for MSP show differences in prototype scores for similar loss values, highlighting that explanation quality can differ along different training runs even if it results in the same loss. This indicates that improvements in model performance are not always associated with better prototype explanations. In contrast, some models can optimize specific aspects of explanation quality such as Continuity and Covariate Complexity without necessarily achieving the lowest loss. These observations suggest that ProtoScore provides more detailed insights than relying solely on loss metrics alone. While loss values are useful for evaluating overall reconstruction or prediction performance, they fail to capture the nuances of explanation quality. ProtoScore provides enhanced insights by examining multiple criteria that reflect different dimensions of interpretability. This enables a more comprehensive evaluation, revealing balances and inconsistencies that loss metrics alone may overlook.

Notably, trade-offs exist between certain properties as models are optimized, which can be seen in Appendix~\ref{chap:appendixresults} Table~\ref{tab:results2}. 
For example, properties such as Continuity and Contrastivity tend to correlate. High Contrastivity, which implies significant distances between prototypes, reduces the likelihood of shifts in prototype assignments when input noise is introduced. Additionally, Input-Completness is often accompanied by Confidence. This is evidenced by shorter distances between data points and their respective prototypes when the prototypes are situated near their clusters. On the other hand, Cohesion of Latent Space can influence Input Completeness negatively (as shown in the Wafer dataset using MAP). A dense latent space hinders a prototype's ability to represent clusters due to stricter acceptance thresholds based on intra-cluster distances. These observations underscore the nuanced interplay between metrics, highlighting trade-offs and complementarities that can guide explanation refinement efforts.

\subsection{Outlier Analysis}
This outlier analysis explores the impact of introducing outliers to the SAWSINE dataset, where 3\% of the data points are modified by adding noise. The focus is on understanding how these alterations affect the metrics. The results are shown in Table~\ref{tab:results3} in Appendix~\ref{chap:appendixresults} for the unmodified SAWSINE dataset, the evaluation of the modified dataset with original models, and new models trained on the modified dataset.
Plot~\ref{fig:SAWSINE_outlier_latentspace} in Appendix~\ref{chap:appendixresults} shows the latent space of the MAP model with index 3, evaluated on the dataset both with and without outliers. Through dimension reduction, the plot with outliers appears distorted compared to the original one. However, the four core clusters (two orange and two blue) remain visible, with several points outside the clusters and overlapping due to the altered perspective.
Upon validating the original models with the outlier dataset, Correctness decreases by up to 3 percentage points, due to increased prediction difficulty from noise disrupting natural data patterns. Metrics like Consistency, Contrastivity, and Compactness remain unchanged due to the lack of different training scenarios. Continuity decreases by up to 3 percentage points, reflecting challenges in maintaining smooth prediction transitions with outliers. Confidence worsens by a maximum of 3 percentage points, and Input Completeness declines for one MSP model. Covariate Complexity remains stable for MAP models but varies for MSP models, with some improving and others declining. This corresponds to variation in the Cohesion of Latent Space for MSP models. For MAP models, the Cohesion of Latent Space declines due to data points lying outside clusters.

When validating newly trained models using the outlier dataset, while hyperparameters were selected as before for models without outliers, Correctness generally decreased for MAP models but remained stable for MSP models. Consistency improved slightly compared to mixed validation, though MSP showed significant variability across different models, as seen in other datasets. Continuity worsened for MAP models but was similar to mixed models, with a slight improvement for MSP models. Contrastivity generally improved, while Covariate Complexity and Confidence declined. Input Completeness mostly remained unchanged across models, and latent space cohesion deteriorated for MAP models but varied for MSP models.

Overall, the variability in model performance, influenced by training conditions, is evident, with MSP models exhibiting more variability than MAP models. This is consistent with previous findings across various datasets, emphasizing the diverse training influences on MSP models and the observed variations in multiple models for the same dataset. The outlier analysis results supports the benchmark tool's correct and effective functioning, as it responds in general predictably when outliers are introduced into the dataset. The changes in scores are expected when certain points in the dataset are more distant from distinct clusters.

\section{Discussion}
\label{chap:discussion}
\subsection{Reproducibility}
A key advantage of ProtoScore is its reproducibility, as most metrics are deterministic, ensuring consistent benchmark results across runs. Only the Continuity metric fluctuates due to the introduction of stochastic noise. In theory, with an infinite number of samples, Continuity converges. However, with the sample number used in this benchmark, only minor fluctuations are observed, leading to scores that lie within a range of minor uncertainties.

\subsection{Challenges for Interpretation}
ProtoScore can be challenging to interpret for underfitted models due to clustering issues that affect the calculated metrics. Underfitted models struggle to accurately capture underlying patterns and to correctly display the data in the latent space in distinct clusters. This is reflected in a low Cohesion of Latent Space score but is not always displayed in the other criteria or even leads to misinterpretable improvements in these criteria. For example, the dependence on clustering can obscure insights into Input Completeness, as the threshold for a representative prototype shrinks. For underfitted models, some scores lead to confusion about the model's overall performance, and thus each criterion must be interpreted in the context of all other criteria to avoid misinterpretations.

Total scores provide a useful summary, but can obscure weaknesses in individual properties. For example, a high total score might mask deficiencies in critical properties such as Correctness that are critical for certain applications. 
Certain properties affect the behavior of other properties, as described in Section~\ref{chap:Experiments}, leading to potential misinterpretations. Understanding these correlations and their underlying principles is vital, as each metric provides complementary insights into explanation quality. Careful consideration must be taken to ensure that these metrics are interpreted in context, particularly when designing or refining models to optimize multiple criteria simultaneously. Moreover, the importance of each property varies across use cases. For instance, in domains such as healthcare or finance, Correctness and Consistency are crucial. In contrast, exploratory tasks may prioritize Contrastivity for better visible class distinction. Additionally, results from one dataset might not generalize to others with different characteristics. Properties such as Consistency and Continuity can behave differently based on dataset complexity and variability.

\subsection{Supporting Model Selection}
Benchmarking multiple models, as discussed in Section~\ref{chap:Experiments}, offers valuable insights into the trade-offs between different methods. The comparison of MAP and MSP reveals similar results across various datasets, highlighting the strengths and weaknesses of each approach. Thus, ProtoScore enables users to select the method that best aligns with their priorities. However, caution is necessary, as several factors can lead to misinterpretations of benchmark results.

The findings in Section~\ref{chap:Experiments} emphasize using ProtoScore as a comprehensive evaluation tool rather than relying solely on loss values or a single criterion. Analyzing the interplay between different properties and considering dataset-specific and method-specific behaviors enables users to make well-informed decisions about model selection and optimization strategies.

\subsection{Quantitative Evaluation Metrics}
This paper underscores the significance of evaluating explainability methods using quantitative metrics, offering an objective approach that reduces reliance on human evaluation. However, developing a comprehensive computational benchmark applicable across all methods poses significant challenges. Thus, we focus on prototype-based methods and use the Co-12 properties as evaluation criteria. While these criteria can be applied to other methods, direct comparisons to our prototype metrics are difficult due to \ac{xai} approach specific definitions, which complicate assessments even when targeting the same Co-12 property.
Moreover, interpretability remains an inherently subjective concept, influenced by the context of explanations and the backgrounds of both providers and receivers. We advocate for quantitative metrics to enhance explanations before testing them in user studies. As our understanding of human interpretability in \ac{xai} progresses, these metrics may replace qualitative assessments. As noted by \citet{VEERAPPA2022101539}, \ac{xai} evaluation should be an interdisciplinary effort that integrates both human and computational elements. Our library is designed to expedite the processes of development, refinement, and failure identification prior to conducting user studies.

\section{Conclusion and Future Work}
\label{chap:conclusion}
In conclusion, this research establishes a crucial framework for evaluating prototype-based \acf{xai} methods, particularly in the context of different data types. By integrating the Co-12 properties in an applicable and quantitative way, our proposed ProtoScore framework provides a comprehensive metric system that allows for effective and automated assessment of prototype quality across various data types. Through practical applications and detailed analyses, we demonstrate how ProtoScore can guide practitioners in selecting appropriate \ac{xai} methods, while minimizing the costs associated with user studies. Our findings underscore the importance of reliable and reproducible evaluation methods to enhance the use of explainability methods. 

%\section{Outlook on Future Work}
Future work should aim for refining these metrics and be completed by user-centric evaluations to enhance the practical applicability of \ac{xai} methods. It is crucial to extend metrics to include not only latent space considerations, but also further parts of the model, and develop a more comprehensive evaluation framework that connects quantitative metrics with human-centric interpretability assessments. With regard to synthetic data, while such datasets can serve as additional resources to assess the plausibility of explanations, they should not replace real-world tests. Given our frequent lack of knowledge about the ground truth generative process for real datasets, integrating synthetic datasets with known clusters and prototypes could facilitate future analyses. However, this integration must balance the realism of the data with the availability of ground truth.
Another enhancement could include dynamic weighting of properties based on user-defined priorities, improved visualizations of property interdependencies, and mechanisms to adaptively optimize models for specific property combinations.
Moreover, we neglected time series-specific metrics and only used the Euclidean distance since our goal for the framework is to be versatile across different data types, including both time series and image data. A valuable extension of our framework is to allow users to choose from a range of metrics tailored to their specific data type and requirements.
Since the best total score in our analysis was $0.76$ using MAP on the SAWSINE and Wafer dataset, there is also a need for further development across prototype methods to achieve a better prototype quality.


\begin{acks}
Certain sections of this paper were developed using a company-specific version of ChatGPT (4o mini). It assisted in generating LaTeX code for tables and enhancing the drafts of specific sections.
We thank our colleagues and the reviewers of the FAccT 2025 conference, who
helped to improve this paper with their valuable feedback. This paper is funded in parts by the German Research Foundation (DFG) Priority Programme (SPP 2422) {\textquotedblleft{Data-Driven Process Modelling in Forming Technology}\textquotedblright} (Subproject {\textquotedblleft{Transparent AI-supported process modeling in drop forging}\textquotedblright} 520195047), by the Fraunhofer Gesellschaft under grant no. PREPARE 40-02702 (project {\textquotedblleft{ML4Safety}\textquotedblright}) and by the Federal Ministry for Economic Affairs and Climate Action under the Industrial Collective Research (IGF) program on the basis of a decision by the German Bundestag (IGF project 22929 BG of the Federation of Quality Research and Science (FQS) {\textquotedblleft{AIQualify - Framework for Qualifying AI Systems in Industrial Quality Inspection}\textquotedblright}).
\end{acks}


\bibliographystyle{ACM-Reference-Format}
\bibliography{references}

@InProceedings{Nauta.2023b,
author="Nauta, Meike
and Seifert, Christin",
title="The Co-12 Recipe for Evaluating Interpretable Part-Prototype Image Classifiers",
booktitle="Explainable Artificial Intelligence",
editor="Longo, Luca",
year="2023",
publisher="Springer Nature Switzerland",
journal=" Communications in Computer and Information Science Explainable Artificial Intelligence",
address="Cham.",
pages="397--420",
volume="1901",
isbn="978-3-031-44064-9",
url="https://doi.org/10.1007/978-3-031-44064-9_21"             
}

@article{Nauta.2023,
    author = {Nauta, Meike and Trienes, Jan and Pathak, Shreyasi and Nguyen, Elisa and Peters, Michelle and Schmitt, Yasmin and Schl\"{o}tterer, J\"{o}rg and van Keulen, Maurice and Seifert, Christin},
    title = {From Anecdotal Evidence to Quantitative Evaluation Methods: A Systematic Review on Evaluating Explainable AI},
    year = {2023},
    issue_date = {December 2023},
    publisher = {Association for Computing Machinery},
    address = {New York, NY, USA},
    volume = {55},
    number = {13s},
    issn = {0360-0300},
    url = {https://doi.org/10.1145/3583558}                 ,  
    doi = {10.1145/3583558},
    journal = {ACM Computing Surveys},
    month = jul,
    articleno = {295},
    numpages = {42},
    }

@article{dzindolet2003,
title = {The role of trust in automation reliance},
journal = {International Journal of Human-Computer Studies},
volume = {58},
number = {6},
pages = {697-718},
year = {2003},
issn = {1071-5819},
doi = {https://doi.org/10.1016/S1071-5819(03)00038-7}                 ,
url = {https://www.sciencedirect.com/science/article/pii/S1071581903000387},
author = {Mary T. Dzindolet and Scott A. Peterson and Regina A. Pomranky and Linda G. Pierce and Hall P. Beck},
}

@misc{Belaid2022DoWN,
      title={Do We Need Another Explainable AI Method? Toward Unifying Post-hoc XAI Evaluation Methods into an Interactive and Multi-dimensional Benchmark}, 
      author={Mohamed Karim Belaid and Eyke Hüllermeier and Maximilian Rabus and Ralf Krestel},
      year={2022},
      eprint={2207.14160},
      url={https://arxiv.org/abs/2207.14160}, 
}

@article{Theissler2022ExplainableAF,
 author = {Theissler, Andreas and Spinnato, Francesco and Schlegel, Udo and Guidotti, Riccardo},
 year = {2022},
 title = {Explainable AI for Time Series Classification: A Review, Taxonomy and Research Directions},
 pages = {100700--100724},
 volume = {10},
 journal = {IEEE Access},
 doi = {10.1109/ACCESS.2022.3207765},
}

@misc{Sattarzadeh2021SVEAAS,
 author={Sattarzadeh, Sam and Sudhakar, Mahesh and Plataniotis, Konstantinos N.},
  journal={2021 IEEE/CVF International Conference on Computer Vision Workshops (ICCVW)}, 
  title={SVEA: A Small-scale Benchmark for Validating the Usability of Post-hoc Explainable AI Solutions in Image and Signal Recognition}, 
  year={2021},
  volume={},
  number={},
  pages={4141-4150},
  doi={10.1109/ICCVW54120.2021.00462}}

@inproceedings{mohseni2020quantitative,
author = {Mohseni, Sina and Block, Jeremy E and Ragan, Eric},
title = {Quantitative Evaluation of Machine Learning Explanations: A Human-Grounded Benchmark},
year = {2021},
isbn = {9781450380171},
publisher = {Association for Computing Machinery},
address = {New York, NY, USA},
url = {https://doi.org/10.1145/3397481.3450689}                      ,
doi = {10.1145/3397481.3450689},
pages = {22–31},
numpages = {10},
booktitle = {Proceedings of the 26th International Conference on Intelligent User Interfaces},
location = {College Station, TX, USA},
series = {IUI '21}
}

@article{liu2021synthetic,
author = {Liu, Yang and Khandagale, Sujay and Khandagale, Sujay and White, Colin and Neiswanger, Willie},
 journal = {Proceedings of the Neural Information Processing Systems Track on Datasets and Benchmarks},
 editor = {J. Vanschoren and S. Yeung},
 pages = {},
 title = {Synthetic Benchmarks for Scientific Research in Explainable Machine Learning},
 url = {https://datasets-benchmarks-proceedings.neurips.cc/paper_files/paper/2021/file/c16a5320fa475530d9583c34fd356ef5-Paper-round2.pdf},
 volume = {1},
 year = {2021}
}

@misc{Le2023BenchmarkingEA,
  title     = {Benchmarking eXplainable AI - A Survey on Available Toolkits and Open Challenges},
  author    = {Le, Phuong Quynh and Nauta, Meike and Nguyen, Van Bach and Pathak, Shreyasi and Schlötterer, Jörg and Seifert, Christin},
  journal = {Proceedings of the Thirty-Second International Joint Conference on
               Artificial Intelligence, {IJCAI-23} Survey Track},
  publisher = {International Joint Conferences on Artificial Intelligence Organization},
  editor    = {Edith Elkind},
  pages     = {6665--6673},
  year      = {2023},
  month     = {8},
  doi       = {10.24963/ijcai.2023/747},
  url       = {https://doi.org/10.24963/ijcai.2023/747}                       ,
}

@misc{Napolitano2024BONESAB,
  title={BONES: a benchmark for neural estimation of shapley values},
  author={Davide Napolitano and Luca Cagliero},
  journal={2024 IEEE 18th International Conference on Application of Information and Communication Technologies (AICT)},
  year={2024},
  pages={1-6},
doi = {10.1109/AICT61888.2024.10740433 }
}

@inproceedings{Sithakoul2024BEExAIBT,
author="Sithakoul, Samuel
and Meftah, Sara
and Feutry, Cl{\'e}ment",
editor="Longo, Luca
and Lapuschkin, Sebastian
and Seifert, Christin",
title="BEExAI: Benchmark to Evaluate Explainable AI",
booktitle="Explainable Artificial Intelligence",
year="2024",
publisher="Springer Nature Switzerland",
journal="Communications in Computer and Information Science",
volume="2153",
address="Cham.",
pages="445--468",
isbn="978-3-031-63787-2",
url= "https://doi.org/10.1007/978-3-031-63787-2_23"                
}

@article{ROUSSEEUW198753,
title = {Silhouettes: A graphical aid to the interpretation and validation of cluster analysis},
journal = {Journal of Computational and Applied Mathematics},
volume = {20},
pages = {53-65},
year = {1987},
doi = {https://doi.org/10.1016/0377-0427(87)90125-7}       ,
author = {Peter J. Rousseeuw},
}

@inproceedings{Schlegel19,
author = { Schlegel, Udo and Arnout, Hiba and El-Assady, Mennatallah and Oelke, Daniela and Keim, Daniel A. },
booktitle = { 2019 IEEE/CVF International Conference on Computer Vision Workshop (ICCVW) },
title = {{ Towards A Rigorous Evaluation Of XAI Methods On Time Series }},
year = {2019},
volume = {},
ISSN = {},
pages = {4197-4201},
doi = {10.1109/ICCVW.2019.00516},
url = {https://doi.ieeecomputersociety.org/10.1109/ICCVW.2019.00516}                       ,
publisher = {IEEE Computer Society},
address = {Los Alamitos, CA, USA},
month ={Oct}
}

@misc{narayanan2024prototypebased,
title={Prototype-Based Methods in Explainable {AI} and Emerging Opportunities in the Geosciences},
author={Anushka Narayanan and Karianne Bergen},
booktitle={Proceedings of the 41 st International Conference on Machine
Learning (ICML)},
year={2024},
publisher = {AI for Science Workshop},
address={Vienna, Austria},
doi = {10.48550/arXiv.2410.19856 }
}

@article{VEERAPPA2022101539,
title = {Validation of XAI explanations for multivariate time series classification in the maritime domain},
journal = {Journal of Computational Science},
volume = {58},
pages = {101539},
year = {2022},
issn = {1877-7503},
doi = {https://doi.org/10.1016/j.jocs.2021.101539}       ,
author = {Manjunatha Veerappa and Mathias Anneken and Nadia Burkart and Marco F. Huber},
}

@article{quantus,
 author = {Anna Hedström and Leander Weber and Dilyara Bareeva and Daniel Krakowczyk and Franz Motzkus and Wojciech Samek and Sebastian Lapuschkin and Marina M.-C. Höhne},
  title = {Quantus: An Explainable AI Toolkit for Responsible Evaluation of Neural Network Explanations and Beyond},
  journal = {Journal of Machine Learning Research},
  volume = {24},
  pages = {1--11},
  year = {2023},
url = {http://jmlr.org/papers/v24/22-0142.html},
number  = {34},
}

@misc{kokhlikyan2020captumunifiedgenericmodel,
      title={Captum: A unified and generic model interpretability library for PyTorch}, 
      author={Narine Kokhlikyan and Vivek Miglani and Miguel Martin and Edward Wang and Bilal Alsallakh and Jonathan Reynolds and Alexander Melnikov and Natalia Kliushkina and Carlos Araya and Siqi Yan and Orion Reblitz-Richardson},
      year={2020},
      eprint={2009.07896},
      journal={arXiv},
      url={https://arxiv.org/abs/2009.07896}, 
}

@misc{arya2021aiexplainability360impact,
      title={AI Explainability 360: Impact and Design}, 
      author={Vijay Arya and Rachel K. E. Bellamy and Pin-Yu Chen and Amit Dhurandhar and Michael Hind and Samuel C. Hoffman and Stephanie Houde and Q. Vera Liao and Ronny Luss and Aleksandra Mojsilovic and Sami Mourad and Pablo Pedemonte and Ramya Raghavendra and John Richards and Prasanna Sattigeri and Karthikeyan Shanmugam and Moninder Singh and Kush R. Varshney and Dennis Wei and Yunfeng Zhang},
      year={2021},
      eprint={2109.12151},
      journal={arXiv},
      url={https://arxiv.org/abs/2109.12151}, 
}

@inproceedings{Agarwal22,
author = {Agarwal, Chirag and Krishna, Satyapriya and Saxena, Eshika and Pawelczyk, Martin and Johnson, Nari and Puri, Isha and Zitnik, Marinka and Lakkaraju, Himabindu},
title = {OpenXAI: towards a transparent evaluation of post hoc model explanations},
year = {2024},
isbn = {9781713871088},
publisher = {Curran Associates Inc.},
address = {Red Hook, NY, USA},
booktitle = {Proceedings of the 36th International Conference on Neural Information Processing Systems},
articleno = {1148},
numpages = {16},
location = {New Orleans, LA, USA},
series = {NIPS '22},
url = {https://dl.acm.org/doi/10.5555/3600270.3601418       }
}

@book{Molnar.2022,
 author = {Molnar, Christoph},
 year = {2022},
 title = {Interpretable machine learning: A guide for making black box models explainable},
 url = {https://christophm.github.io/interpretable-ml-book/},
 address = {Munich, Germany},
 edition = {Second edition},
 publisher = {{Christoph Molnar}}
}

@misc{wang2023learningsupporttrivialprototypes,
      title={Learning Support and Trivial Prototypes for Interpretable Image Classification}, 
      author={Chong Wang and Yuyuan Liu and Yuanhong Chen and Fengbei Liu and Yu Tian and Davis J. McCarthy and Helen Frazer and Gustavo Carneiro},
      year={2023},
      eprint={2301.04011},
      journal={arXiv},
      url={https://arxiv.org/abs/2301.04011}, 
}

@misc{Hoffman2018,
      title={Metrics for Explainable AI: Challenges and Prospects}, 
      author={Robert R. Hoffman and Shane T. Mueller and Gary Klein and Jordan Litman},
      year={2019},
      eprint={1812.04608},
      journal={arXiv},
      url={https://arxiv.org/abs/1812.04608}, 
}

@inproceedings{Wang.19,
author = {Wang, Danding and Yang, Qian and Abdul, Ashraf and Lim, Brian Y.},
title = {Designing Theory-Driven User-Centric Explainable AI},
year = {2019},
isbn = {9781450359702},
publisher = {Association for Computing Machinery},
address = {New York, NY, USA},
url = {https://doi.org/10.1145/3290605.3300831                                         } ,
doi = {10.1145/3290605.3300831 },
booktitle = {Proceedings of the 2019 CHI Conference on Human Factors in Computing Systems},
pages = {1–15},
numpages = {15},
location = {Glasgow, Scotland Uk},
series = {CHI '19}
}

@inproceedings{hase-bansal-2020-evaluating,
    title = "Evaluating Explainable {AI}: Which Algorithmic Explanations Help Users Predict Model Behavior?",
    author = "Hase, Peter  and
      Bansal, Mohit",
    editor = "Jurafsky, Dan  and
      Chai, Joyce  and
      Schluter, Natalie  and
      Tetreault, Joel",
    booktitle = "Proceedings of the 58th Annual Meeting of the Association for Computational Linguistics",
    month = jul,
    year = "2020",
    address = "Online",
    publisher = "Association for Computational Linguistics",
    url = "https://aclanthology.org/2020.acl-main.491",
    doi = "10.18653/v1/2020.acl-main.491",
    pages = "5540--5552",
}

@article{pmlr-v189-obermair23a,
  title = 	 {Example or Prototype? Learning Concept-Based
 Explanations in Time-Series},
  author =       {Obermair, Christoph and Fuchs, Alexander and Pernkopf, Franz and Felsberger, Lukas and Apollonio, Andrea and Wollmann, Daniel},
  journal = 	 {Proceedings of The 14th Asian Conference on Machine
 Learning},
  pages = 	 {816--831},
  year = 	 {2023},
  editor = 	 {Khan, Emtiyaz and Gonen, Mehmet},
  volume = 	 {189},
  month = 	 {12--14 Dec},
  publisher =    {PMLR},
  url = 	 {https://proceedings.mlr.press/v189/obermair23a.html},
}

@article{McGuirl2006SupportingTC,
  title={Supporting Trust Calibration and the Effective Use of Decision Aids by Presenting Dynamic System Confidence Information},
  author={John M. McGuirl and Nadine B. Sarter},
  journal={Human Factors: The Journal of Human Factors and Ergonomics Society},
  year={2006},
  volume={48},
  pages={656 - 665},
doi = {10.1518/001872006779166334},
}

@article{Rudin19,
author = {Rudin, Cynthia},
year = {2019},
month = {05},
pages = {206--215},
title = {Stop Explaining Black Box Machine Learning Models for High Stakes Decisions and Use Interpretable Models Instead},
volume = {1},
journal = {Nature Machine Intelligence},
doi = {10.1038/s42256-019-0048-x}
}

@misc{Li_Liu_Chen_Rudin_2018, 
title={Deep Learning for Case-Based Reasoning Through Prototypes: A Neural Network That Explains Its Predictions}, 
volume={32},
url={https://ojs.aaai.org/index.php/AAAI/article/view/11771}, DOI={10.1609/aaai.v32i1.11771 }, 
number={1}, 
journal={Proceedings of the AAAI Conference on Artificial Intelligence}, author={Li, Oscar and Liu, Hao and Chen, Chaofan and Rudin, Cynthia}, year={2018}, 
month={Apr.}
}

@misc{Tripathy.2022,
 author = {Tripathy, Sarthak Manas and Chouhan, Ashish and Dix, Marcel and Kotriwala, Arzam and Klopper, Benjamin and Prabhune, Ajinkya},
 title = {Explaining Anomalies in Industrial Multivariate Time-series Data with the help of eXplainable AI},
 pages = {226--233},
 publisher = {IEEE},
 isbn = {978-1-6654-2197-3},
 journal = {2022 IEEE International Conference on Big Data and Smart Computing (BigComp)},
 year = {2022},
 doi = {10.1109/BigComp54360.2022.00051},
}

@article{Weber24,
author = {Weber, Rosina O and Johs, Adam J and Goel, Prateek and Silva, Jo\~{a}o Marques},
title = {XAI is in trouble},
year = {2024},
issue_date = {Fall 2024},
publisher = {John Wiley \& Sons, Inc.},
address = {USA},
volume = {45},
number = {3},
issn = {0738-4602},
url = {https://doi.org/10.1002/aaai.12184                                         } ,
doi = {10.1002/aaai.12184 },
journal = {AI Mag.},
month = jul,
pages = {300–316},
numpages = {17}
}

@article{gee2019explaining,
  title={Explaining Deep Classification of Time-Series Data with Learned Prototypes},
  author={Gee, Alan H and Garcia-Olano, Diego and Ghosh, Joydeep and Paydarfar, David},
  journal={CEUR workshop proceedings},
  volume={2429},
  pages={15--22},
  year={2019},
  organization={NIH Public Access},
editor = {Wiratunga, Nirmalie and Coenen, Frans and Sani, Sadiq},
  url = {http://dblp.uni-trier.de/db/conf/ijcai/khd2019.html#GeeGGP19},
}

@misc{WaferDataset,
  title = {Wafer Dataset - UCR Time Series Classification Repository},
  author = {Robert Thomas Olszewski},
  year = {2001},
  note = {Accessed: 2024-10-20},
  url = {https://timeseriesclassification.com/description.php?Dataset=Wafer}
}

@misc{ECG200,
  title = {ECG200 Data Set},
  author = {Robert Thomas Olszewski},
  year = {2001},
  note = {Accessed: 2024-10-20},
  url = {https://www.timeseriesclassification.com/description.php?Dataset=ECG200}
}

@misc{olszewski2001,
  title={Generalized Feature Extraction for Structural Pattern Recognition in Time-Series Data},
  author={Olszewski, Robert Thomas},
  year={2001},
  publisher={Carnegie Mellon University}
}

@misc{FordA,
  title = {FordA Data Set},
  author = {Bagnall, Anthony},
  year = {2008},
  url = {https://timeseriesclassification.com/description.php?Dataset=Forda},
  note = {Accessed: 2024-10-20}
}

@misc{FordB,
  title = {FordB Data Set},
  author = {Bagnall, Anthony},
  year = {2008},
  url = {https://timeseriesclassification.com/description.php?Dataset=FordB},
  note = {Accessed: 2024-10-20}
}

@misc{StarLight,
  title = {StarLightCurves},
  author = {E. Keogh, L. Wei},
  url = {http://www.timeseriesclassification.com/description.php?Dataset=StarlightCurves},
  note = {Accessed: 2025-04-10},
year = {2009}
}

@misc{umap,
      title={UMAP: Uniform Manifold Approximation and Projection for Dimension Reduction}, 
      author={Leland McInnes and John Healy and James Melville},
      year={2020},
      eprint={1802.03426},
      journal={arXiv},
      url={https://arxiv.org/abs/1802.03426}, 
}

@misc{
hanawa2021evaluation,
title={Evaluation of Similarity-based Explanations},
author={Kazuaki Hanawa and Sho Yokoi and Satoshi Hara and Kentaro Inui},
booktitle={International Conference on Learning Representations},
year={2021},
}

@misc{NIPS2016_5680522b,
 author = {Kim, Been and Khanna, Rajiv and Koyejo, Oluwasanmi O},
 booktitle = {Advances in Neural Information Processing Systems},
 editor = {D. Lee and M. Sugiyama and U. Luxburg and I. Guyon and R. Garnett},
 pages = {},
 publisher = {Curran Associates, Inc.},
 title = {Examples are not enough, learn to criticize! Criticism for Interpretability},
 url = {https://proceedings.neurips.cc/paper_files/paper/2016/file/5680522b8e2bb01943234bce7bf84534-Paper.pdf},
 volume = {29},
 year = {2016}
}

@inproceedings{Fraser22,
author = {Fraser, Henry and Simcock, Rhyle and Snoswell, Aaron J.},
title = {AI Opacity and Explainability in Tort Litigation},
year = {2022},
isbn = {9781450393522},
publisher = {Association for Computing Machinery},
address = {New York, NY, USA},
url = {https://doi.org/10.1145/3531146.3533084                        },
booktitle = {Proceedings of the 2022 ACM Conference on Fairness, Accountability, and Transparency},
pages = {185–196},
numpages = {12},
location = {Seoul, Republic of Korea},
series = {FAccT '22}
}

@misc{EuropeanParliament2016a,
  date       = {2016-05-04},
  location   = {OJ L 119, 4.5.2016, p. 1--88},
  title      = {Regulation (EU) 2016/679 of the European Parliament and of the Council},
  url        = {https://data.europa.eu/eli/reg/2016/679/oj},
  titleaddon = {of 27 April 2016 on the protection of natural persons with regard to the processing of personal data and on the free movement of such data, and repealing Directive 95/46/EC (General Data Protection Regulation)},
  author     = {European Parliament and Council of the European Union},
  urldate    = {2025-04-23},
year = {2016}
}

@misc{EuropeanParliament2024,
  date       = {2024-06-13},
  location   = {OJ L 119, 4.5.2016, p. 1--88},
  title      = {Regulation (EU) 2024/1689 of the European Parliament and of the Council},
  url        = {http://data.europa.eu/eli/reg/2024/1689/oj},
  titleaddon = {of 13 June 2024 laying down harmonised rules on artificial intelligence and amending Regulations (EC) No 300/2008, (EU) No 167/2013, (EU) No 168/2013, (EU) 2018/858, (EU) 2018/1139 and (EU) 2019/2144 and Directives 2014/90/EU, (EU) 2016/797 and (EU) 2020/1828 (Artificial Intelligence Act)},
  author     = {European Parliament and Council of the European Union},
  urldate    = {2025-04-23},
year = {2024}
}

@misc{CJEU2023,
author = {CJEU},
year = {07.12.2023},
title = {SCHUFA Holding (Scoring)},
note = {Judgement, C‑634/21, ECLI EU:C:2023:957}
}

@misc{CJEU2025,
author = {CJEU},
year = {27.02.2025},
title = {Dun \& Bradstreet Austria},
note = {Judgement, C‑203/22, ECLI EU:C:2025:117}
}

\clearpage
\appendix
\section{Existing Evaluation Frameworks}
\label{chap:evaluationframeworks}
\emph{Captum}~\cite{kokhlikyan2020captumunifiedgenericmodel} offers two evaluation metrics—infidelity and sensitivity—across 22 attribution methods applicable to various data formats, including images, tabular data, and text. However, the primary focus of this library is on providing multiple implementations of \ac{xai} methods rather than establishing a comprehensive benchmark.

\emph{AIX360}~\cite{arya2021aiexplainability360impact} also addresses time series data, offering methods for both global and local explanations, as well as post-hoc, ante-hoc, and data-centered \ac{xai} approaches. Nonetheless, it includes only two evaluation metrics: faithfulness and monotonicity, and is primarily designed for \ac{xai} application rather than evaluation, excluding prototype-based methods.
XAI-Bench~\cite{liu2021synthetic} evaluates \ac{xai} algorithms based on five metrics: Faithfulness, Monotonicity, ROAR, GT-Shapley, and Infidelity, using synthetic data. This framework assesses six \ac{xai} methods: SHAP, LIME, MAPLE, SHAPR, L2X, and breakDown.

\emph{OpenXAI}~\cite{Agarwal22} builds upon the widely-used metrics from Quantus by introducing additional metrics to assess the fairness of explanations. The library also includes synthetic data generation tools, facilitating comprehensive ground truth comparisons, and features leaderboards for six selected tabular datasets, enabling comparative evaluations of Neural Network and Logistic Regression models.
\emph{Compare-xAI}~\cite{Belaid2022DoWN} serves as a quantitative benchmark tool designed to evaluate and compare \ac{xai} algorithms based on functional testing methods. These tests are categorized into five groups: fidelity, fragility, stability, simplicity, and stress. The benchmark employs a hierarchical scoring system to provide a comprehensibility score, allowing users to evaluate algorithms based on their specific needs and levels of expertise. However, this tool is limited to \ac{xai} methods that provide feature importance (global explanations).

The \emph{SVEA} benchmark~\cite{Sattarzadeh2021SVEAAS} aims to evaluate various attribution methods, including Grad-CAM, Integrated Gradients, and RISE, across different metrics. It enables researchers to assess the usability of attribution methods based on properties such as soundness, Completeness, contextfulness, and actionability, without imposing high computational demands by using the small MNIST-1D dataset.

\emph{BEExAI}~\cite{Sithakoul2024BEExAIBT} is proposed as a benchmark tool facilitating large-scale comparisons of different post-hoc \ac{xai} methods through a selected set of evaluation metrics. This library calculates nine evaluation metrics focused on three core properties of explainability: Faithfulness, Robustness, and Complexity, and is designed to evaluate eight explainability methods, including LIME, Shapley Values, and Integrated Gradients, but only on tabular datasets.

\emph{BONES}~\cite{Napolitano2024BONESAB} is a benchmark tool for neural estimation of Shapley Values. It provides researchers with a suite of state-of-the-art neural and traditional Shapley value estimators, such as KernelSHAP, FastSHAP, and DASP, along with a selection of commonly used benchmark datasets for tabular and image data. The library includes ante-hoc modules for train black-box models and specific functions for computing popular evaluation metrics, such as L1, L2 distances, and AUC for image data, as well as visualizing results. It also features functions for quantifying estimation errors, assessing computational costs, and comparing different explainers.

\section{Experiments}
\label{chap:datasetsmethods}
\subsection{Used Datasets}
The FordA dataset~\cite{FordA} is designed for fault detection in Ford's powertrain systems and contains sensor data collected from a Ford vehicle's powertrain system under various operational conditions. It is commonly used for time series classification tasks and contains $3,601$ train instances and $1,320$ test instances. Each time series corresponds to a measurement of engine noise captured by a motor sensor. For FordA the train and test dataset were collected in typical operating conditions, with minimal noise contamination.

The dataset FordB~\cite{FordB} is similar to FordA. This dataset contains $3,636$ training instances and $810$ test samples, each sample consists of $500$ measurements of engine noise and a classification. For FordB the train data was collected in typical operating conditions, but the test data samples were collected under noisy conditions.

The Wafer dataset~\cite{WaferDataset} is used for time series classification in semiconductor manufacturing, specifically for identifying conditions or faults in wafers. The dataset features two classes: normal and abnormal. The dataset contains $1,000$ train instances and $6,164$ test samples. However, there is a significant class imbalance, with only $10.7~\%$ of the training data and $12.1~\%$ of the test data classified as abnormal.

The ECG200 dataset~\cite{ECG200} consists of measurements of cardiac electrical activity recorded from electrodes placed at various locations on the body. Each dataset contains recordings from a single electrode during one heartbeat. The ECG dataset comprises a total of $200$ datasets, of which $133$ are identified as normal and $67$ as abnormal. This classification is crucial for detecting conditions such as bradycardia, which is characterized by a slower than normal heart rate. The dataset is part of a larger effort to leverage deep learning techniques to improve the detection and interpretation of critical cardiac events, particularly in high-risk populations such as preterm infants.
Both the Wafer dataset and ECG200 dataset were formatted by \citet{olszewski2001}.

The SAWSINE dataset was created by \citet{pmlr-v189-obermair23a} to simulate signals from machine sensors operating in a noisy environment. This dataset comprises four basic time series shapes that serve as ground truth signals. To introduce realism and complexity, both multiplicative and additive noise were added to these signals. The noise levels varied, with amplitudes ranging from $0$ to $1.1$, and were drawn from a uniform distribution. The dataset we used has $8,000$ samples divided into two equal classes with $4,000$ and $4,000$ data points.

The StarLightCurve dataset~\cite{StarLight} contains a collection of phase-aligned starlight curves that represents the brightness of a celestial object (e.g., a star) as a function of time. The analysis of these curves is important in astronomy, particularly for studying the variability of celestial sources.
Each starlight curve in this dataset has a length of 1,024 data points, capturing the intensity of light over time. The class labels (3 distinct classes) for each curve have been determined by an expert in the field. 
The dataset has a training set of $1,000$ samples and a test set of $8,236$ samples.
\pagebreak
\subsection{Used Prototype Methods}
\label{chap:appendixmethods}
\emph{MAP} (Model-Agnostic Prototype)~\cite{pmlr-v189-obermair23a} focuses on generating interpretable explanations for time series classification models using a model-agnostic approach based on prototypes derived from an autoencoder. The method employs an autoencoder architecture consisting of an encoder that maps input time series data to a latent space and a decoder that reconstructs the original data. Prototypes are derived from this latent space using k-means clustering, serving as abstract representations of concepts relevant to the data. The train objective optimizes for reconstruction accuracy while also including a similarity loss to ensure diversity among prototypes. This model-agnostic design enables the method to work independently of any specific classification model, allowing it to generate explanations for existing models without modification.

The prototype method \emph{MSP} (Model-Specific Prototype)~\cite{gee2019explaining} is a model specific approach. It integrates an autoencoder architecture to enhance the interpretability of deep learning models applied to time series classification tasks. The model consists of an encoder that transforms input time series data into a lower-dimensional latent space, capturing essential features while minimizing information loss. The decoder then reconstructs the original data from this latent representation. To classify the data, the learned latent representations are fed into a prototype network that identifies and learns multiple prototype vectors corresponding to different classes. A significant advancement in this approach is the introduction of a prototype diversity penalty in the loss function. This penalty encourages the model to learn diverse prototypes, improving their distribution across the latent space and enhancing the model’s ability to classify overlapping classes. The loss function balances several components, including classification loss, reconstruction loss, and the diversity penalty, ensuring that the learned prototypes are both informative and distinct.

\subsection{Architecture and Hyperparameter Used in the Prototype Models}
\label{chap:appendixhyperparameter}
The hyperparameters and resulting loss values for the trained models are presented in this chapter in Table~\ref{tab:hyperparameter}, offering insights into the configurations used during model training.

The architecture used for the MAP models includes an Encoder and Decoder, and a separate trained classifier. The Encoder comprises a Flatten layer that transforms the input data into a one-dimensional vector, making it suitable for processing by subsequent dense layers. It also includes a dense layer that outputs a latent representation with a specified dimension, as detailed in Table~\ref{tab:hyperparameter}. This layer uses an L1 activity regularizer to encourage sparsity in the latent representation, which helps mitigate overfitting.
The Decoder is responsible for reconstructing the original input from the latent representation generated by the Encoder. It consists of a dense layer with $300$ units that initiates the reconstruction process, followed by another dense layer with $300$ units that employs a sigmoid activation function. This function constrains the output values between 0 and 1. The final and third dense layer outputs a vector with dimensions equal to the product of the original input dimensions, effectively reconstructing the data to its original form.
Lastly, the Classifier includes multiple convolutional blocks. Each block applies a 1D convolution with varying filter sizes: the first block employs $128$ filters with a kernel size of $8$, the second block uses $256$ filters with a kernel size of 5, and the third block uses $128$ filters with a kernel size of $3$. Each convolutional layer is followed by batch normalization and ReLU activation to introduce non-linearity. After the convolutional layers, an average pooling layer condenses the feature maps into a single vector, which is then processed by a dense layer with softmax activation to provide class probabilities for classification tasks.

The architecture for the MSP models mirrors that of the MAP models, featuring the same three components. The Encoder is designed as a convolutional network to process input data, starting with the first convolutional block that applies a 1D convolution with $128$ filters and a kernel size of $8$, followed by batch normalization and ReLU activation. The second block uses $256$ filters with a kernel size of $5$, again followed by batch normalization and ReLU activation. The third block consists of $128$ filters with a kernel size of $3$, also followed by batch normalization and ReLU activation. A Global Average Pooling layer condenses the feature maps into a single vector, which is then passed through a dense layer with a specified latent dimension and ReLU activation.
The Decoder for the MSP models reconstructs the original input from the latent representation, which includes a dense layer with $300$ units, followed by another dense layer with $300$ units using a sigmoid activation function to ensure output values fall within the range of $0$ to $1$. The final dense layer outputs a vector with dimensions equal to the product of the original dimensions, effectively reconstructing the input data.
Additionally, the models feature a prototype layer containing a set of randomly initialized, trainable prototype representations, which are used for concept learning based on the latent dimension and number of concepts. The Classifier in the MSP models includes a dense output layer with softmax activation, providing class probabilities.

For both models, the learning rate for the autoencoder is set at $0.0001$, while the classifier uses a learning rate of $0.001$. The autoencoder employs mean squared error (MSE) as the loss function, whereas the classifier uses cross-entropy loss. The models are validated with a separated dataset using also the MSE.

\begin{table*}[b]
    \centering
    \caption{Hyperparameter and loss values used in the prototype models.}
    \label{tab:hyperparameter}
    % Adjust the column definition to add 5 more columns
    \begin{tabular}{c | c c| cc c c c}
        \toprule
        & &  &  & 
           & \textbf{Loss} 
          & \textbf{Loss} & \textbf{Latent Dimension} \\
          \textbf{Index} & \textbf{Dataset} & \textbf{Method} & \textbf{Val Loss} & \textbf{Epochs} 
           & \textbf{Autoencoder} 
          & \textbf{Classifier} & \textbf{ Autoencoder} \\
        \midrule
        
1 & Wafer & MAP &$1.26$ 
            &1 
          &1.52  %1.520292
          & 0.06
          &20 \\
2 & Wafer & MAP &$0.58$
              &3
              &0.71 %0.70963836
              & 0.06
              &20\\

3 & Wafer & MAP &$0.27$
              &15
              &0.55 %0.5530563
              &0.06
              &20\\

4 & Wafer & MAP &$0.28$
              &20
              &0.32 %0.32025546
              &0.06
              &20
              \\

\cellcolor{gray!20}1 & \cellcolor{gray!20}Wafer & \cellcolor{gray!20}MSP & \cellcolor{gray!20}  $0.86$
            & \cellcolor{gray!20}1
            & \cellcolor{gray!20}0.04 %0.04041038
            & \cellcolor{gray!20}0.37 %0.37422925
            &\cellcolor{gray!20}4
            \\

\cellcolor{gray!20}2 & \cellcolor{gray!20}Wafer & \cellcolor{gray!20}MSP & \cellcolor{gray!20}  $0.35$
            & \cellcolor{gray!20}10
            & \cellcolor{gray!20}0.02 %0.018665561
            & \cellcolor{gray!20}0.04 %0.038403712
            &\cellcolor{gray!20}4
            \\
\cellcolor{gray!20}3 & \cellcolor{gray!20}Wafer & \cellcolor{gray!20}MSP & \cellcolor{gray!20}  $0.35$
            & \cellcolor{gray!20}15
            & \cellcolor{gray!20}0.02 %0.01818257603275625
            & \cellcolor{gray!20}0.02 %0.020952407597811726
            &\cellcolor{gray!20}4
            \\

\cellcolor{gray!20}4 & \cellcolor{gray!20}Wafer & \cellcolor{gray!20}MSP & \cellcolor{gray!20}$0.23$
            & \cellcolor{gray!20}30
            & \cellcolor{gray!20}0.01 %0.012731223
            & \cellcolor{gray!20}0.00 %0.0018937078
            &\cellcolor{gray!20}4
            \\

1 & SAWSINE & MAP &$0.37$
            & 1
            &0.49%0.4939649
            &0.01 %0.011668780818581581
            &20
\\

2 & SAWSINE & MAP &$0.18$
            &5
            &0.18 %0.18334438
            &0.01   %0.011668780818581581
            &20
\\

3 & SAWSINE & MAP &$0.17$
            &10 
            &0.17  %0.16782066
            &0.01   %0.011668780818581581
            &20
\\

4 & SAWSINE & MAP &$0.14$
            &30
            &0.15 %0.14862111
            &0.01   %0.011668780818581581
            &20
\\

5 & SAWSINE & MAP &$0.13$
            &50 
            &0.13 %0.1254875
            &0.01   %0.011668780818581581
            &20
\\

\cellcolor{gray!20}1 & \cellcolor{gray!20}SAWSINE & \cellcolor{gray!20}MSP 
            & \cellcolor{gray!20}$0.18$
            & \cellcolor{gray!20}1
            &\cellcolor{gray!20}0.01 %0.009252173
            &\cellcolor{gray!20}0.10 %0.09639229
            &\cellcolor{gray!20}4
\\

\cellcolor{gray!20}2 & \cellcolor{gray!20}SAWSINE & \cellcolor{gray!20}MSP & \cellcolor{gray!20}$0.14$
            & \cellcolor{gray!20}10
            &\cellcolor{gray!20}0.01 %0.007495012
            &\cellcolor{gray!20}0.00 %0.0008665761
            &\cellcolor{gray!20}4
\\

\cellcolor{gray!20}3 & \cellcolor{gray!20}SAWSINE & \cellcolor{gray!20}MSP & \cellcolor{gray!20}$0.14$
            & \cellcolor{gray!20}15
            &\cellcolor{gray!20}0.01 %0.0070156767
            &\cellcolor{gray!20}0.00 %0.00036096753
            &\cellcolor{gray!20}4
\\

\cellcolor{gray!20}4 & \cellcolor{gray!20}SAWSINE & \cellcolor{gray!20}MSP & \cellcolor{gray!20}$0.14$
            & \cellcolor{gray!20}30
            &\cellcolor{gray!20}0.01 %0.007024359
            &\cellcolor{gray!20}0.00 %4.4656415e-05
            &\cellcolor{gray!20}4
            \\
1 & FordB & MAP &$1.02$
            &1
            &1.11 %1.1121053
            & 0.52 %0.5230619311332703
            &20
            \\

2 & FordB & MAP &$0.83$
            &10
            &0.90 %0.8973617
            & 0.52 %0.5230619311332703
            &20
            \\

3 & FordB & MAP &$0.62$
            &30
            &0.72%0.719375
            &0.52 %0.5230619311332703
            &20
            \\

4 & FordB & MAP &$0.49$
            &50
            &0.60 %0.6029627
            & 0.52 %0.5230619311332703
            &20
            \\

\cellcolor{gray!20}1 & \cellcolor{gray!20}FordB & \cellcolor{gray!20}MSP & \cellcolor{gray!20}$1.00$
            & \cellcolor{gray!20}10
            &\cellcolor{gray!20}0.05 %0.04999372
            &\cellcolor{gray!20}0.42%0.42120048
            &\cellcolor{gray!20}4
            \\

\cellcolor{gray!20}2 & \cellcolor{gray!20}FordB & \cellcolor{gray!20}MSP & \cellcolor{gray!20}$1.00$
            & \cellcolor{gray!20}30
            &\cellcolor{gray!20}0.05%0.04998083
            &\cellcolor{gray!20}0.17 %0.17024314
            &\cellcolor{gray!20}4
            \\

\cellcolor{gray!20}3 & \cellcolor{gray!20}FordB & \cellcolor{gray!20}MSP & \cellcolor{gray!20}  $1.00$
            & \cellcolor{gray!20}50
            & \cellcolor{gray!20}0.05 %0.05003784634172916
            &\cellcolor{gray!20}0.22 %0.2212287121348911
            &\cellcolor{gray!20}4
            \\

\cellcolor{gray!20}4 & \cellcolor{gray!20}FordB & \cellcolor{gray!20}MSP & \cellcolor{gray!20}$1.00$
            & \cellcolor{gray!20}100
            &\cellcolor{gray!20}0.05 %0.05026039
            &\cellcolor{gray!20}0.09%0.08797364
            &\cellcolor{gray!20}4
            \\
1 & FordA & MAP &$1.16$
            &1
            &1.23%1.2336476
            &0.22 %0.2207910269498825
            &20
            \\

2 & FordA & MAP &$1.05$
            &5
            &1.02 %1.02189
            &0.22 %0.2207910269498825
            &20\\

3 & FordA & MAP &$0.86$
            &10
            &0.86 %0.86190856
            &0.22 %0.2207910269498825
            &20\\

4 & FordA & MAP &$0.70$
            &30
            &0.70 %0.69704986
            &0.22 %0.2207910269498825
            &20\\

5 & FordA & MAP &$0.58$
            & 50
            &0.56%0.56020904
            & 0.22 %0.2207910269498825
            &20 \\

\cellcolor{gray!20}1 & \cellcolor{gray!20}FordA & \cellcolor{gray!20}MSP & \cellcolor{gray!20}$1.00$
            & \cellcolor{gray!20}10
            & \cellcolor{gray!20}0.05 %0.04989319
            & \cellcolor{gray!20}0.51 %0.51039433
            & \cellcolor{gray!20}4\\

\cellcolor{gray!20}2 & \cellcolor{gray!20}FordA & \cellcolor{gray!20}MSP & \cellcolor{gray!20}$1.00$
            & \cellcolor{gray!20}30
            & \cellcolor{gray!20}0.05%0.049891245 
            & \cellcolor{gray!20}0.31 %0.3079866
            & \cellcolor{gray!20}4
            \\

\cellcolor{gray!20}3 & \cellcolor{gray!20}FordA & \cellcolor{gray!20}MSP & \cellcolor{gray!20}$1.00$
            & \cellcolor{gray!20}100
            & \cellcolor{gray!20}0.05%0.049775366 
            & \cellcolor{gray!20}0.15 %0.14842023
            & \cellcolor{gray!20}4
            \\

\cellcolor{gray!20}4 & \cellcolor{gray!20}FordA & \cellcolor{gray!20}MSP & \cellcolor{gray!20}$0.99$
            & \cellcolor{gray!20}150
            & \cellcolor{gray!20}0.05 %0.049638093
            & \cellcolor{gray!20}0.14%0.13967538
            & \cellcolor{gray!20}4
            \\

1 & ECG200 & MAP &$1.45$
            &3
            &1.52%1.5239303
            & 0.57 %0.566209614276886
            &20 \\

2 & ECG200 & MAP &$0.70$
            &30
            &0.71%0.7096722
            & 0.57 %0.566209614276886
            &20 \\
            
3 & ECG200 & MAP &$0.55$
            &70
            &0.56%0.56449157
            & 0.57 %0.566209614276886
            &20 \\

4 & ECG200 & MAP &   $0.38$
            &200
            &0.40%0.39917916
            & 0.57 %0.566209614276886
            &20 \\
            
5 & ECG200 & MAP &   $0.26$
            & 250
            &0.29%0.28539914
            & 0.57 %0.566209614276886
            &20 \\

\cellcolor{gray!20}1 & \cellcolor{gray!20} ECG200 &\cellcolor{gray!20}MSP & \cellcolor{gray!20}  $1.04$
            & \cellcolor{gray!20}1
            & \cellcolor{gray!20}0.06 %0.05506578367203474
            & \cellcolor{gray!20}0.64 %0.6414018869400024
            & \cellcolor{gray!20}4
            \\

\cellcolor{gray!20}2 & \cellcolor{gray!20}ECG200 & \cellcolor{gray!20}MSP & \cellcolor{gray!20}  $0.78$
            & \cellcolor{gray!20}1
            & \cellcolor{gray!20}0.05%0.044995926
            & \cellcolor{gray!20}0.75%0.74716616
            & \cellcolor{gray!20}4
            \\

\cellcolor{gray!20}3 & \cellcolor{gray!20}ECG200 & \cellcolor{gray!20}MSP & \cellcolor{gray!20}  $0.36$
            & \cellcolor{gray!20}10
            & \cellcolor{gray!20}0.02 %0.017091276
            & \cellcolor{gray!20}1.48%1.4806105
            & \cellcolor{gray!20}4
            \\

\cellcolor{gray!20}4 & \cellcolor{gray!20}ECG200 & \cellcolor{gray!20}MSP & \cellcolor{gray!20}  $0.30$
            & \cellcolor{gray!20}50
            & \cellcolor{gray!20}0.01%.0120046465
            & \cellcolor{gray!20}0.24%0.24065489
            & \cellcolor{gray!20}4
            \\
            
\cellcolor{gray!20}5 & \cellcolor{gray!20}ECG200 & \cellcolor{gray!20}MSP & \cellcolor{gray!20}  $0.29$
            & \cellcolor{gray!20}200
            & \cellcolor{gray!20}0.01 %0.013926975739498934
            & \cellcolor{gray!20}0.12%0.1149315486351649 
            & \cellcolor{gray!20}4
            \\
 1 & StarLightCurve & MAP &$1.07$
            &1
            &1.10
            & 1.30 
            &30 \\

2 & StarLightCurve & MAP &$0.76$
            &5
            &0.79
            & 0.29
            &30 \\
            
3 & StarLightCurve & MAP &$0.55$
            &10
            &0.57
            & 0.27
            &30 \\

4 & StarLightCurve & MAP &   $0.31$
            &30
            &0.32
            & 0.27
            &30 \\
            
5 & StarLightCurve & MAP &   $0.13$
            & 50
            &0.14
            & 0.27
            &30 \\

\cellcolor{gray!20}1 & \cellcolor{gray!20} StarLightCurve &\cellcolor{gray!20}MSP & \cellcolor{gray!20}  $0.42$
            & \cellcolor{gray!20}1
            & \cellcolor{gray!20}0.03
            & \cellcolor{gray!20}0.89
            & \cellcolor{gray!20}6
            \\

\cellcolor{gray!20}2 & \cellcolor{gray!20}StarLightCurve & \cellcolor{gray!20}MSP & \cellcolor{gray!20}  $0.28$
            & \cellcolor{gray!20}5
            & \cellcolor{gray!20}0.01
            & \cellcolor{gray!20}0.50
            & \cellcolor{gray!20}6
            \\

\cellcolor{gray!20}3 & \cellcolor{gray!20}StarLightCurve & \cellcolor{gray!20}MSP & \cellcolor{gray!20}  $0.22$
            & \cellcolor{gray!20}20
            & \cellcolor{gray!20}0.01
            & \cellcolor{gray!20}0.30
            & \cellcolor{gray!20}6
            \\

\cellcolor{gray!20}4 & \cellcolor{gray!20}StarLightCurve & \cellcolor{gray!20}MSP & \cellcolor{gray!20}  $0.20$
            & \cellcolor{gray!20}50
            & \cellcolor{gray!20}0.01
            & \cellcolor{gray!20}0.14
            & \cellcolor{gray!20}6
            \\
            
\cellcolor{gray!20}5 & \cellcolor{gray!20}StarLightCurve & \cellcolor{gray!20}MSP & \cellcolor{gray!20}  $0.12$
            & \cellcolor{gray!20}100
            & \cellcolor{gray!20}0.01
            & \cellcolor{gray!20}0.11
            & \cellcolor{gray!20}6
            \\       
        \bottomrule
    \end{tabular}
\end{table*}

\subsection{Results for Experiments on All Datasets}
\label{chap:appendixresults}
Additional experiments were performed on the Wafer, SAWSINE, FordB, FordA, and StarLightCurves datasets to evaluate our benchmark framework, as presented in Table~\ref{tab:results2}. The ECG200 results from Table~\ref{tab:results} are also included for a better overview. These results evaluate the effectiveness of both the MAP and MSP methods across various metrics, which are further discussed in Section~\ref{chap:Experiments}.

\begin{figure*}[b]
\vspace{1cm}
      \begin{minipage}{0.4\textwidth}
     \centering
        \includegraphics[trim=0cm 0.5cm 0cm 0.2cm, clip, width=1\textwidth]{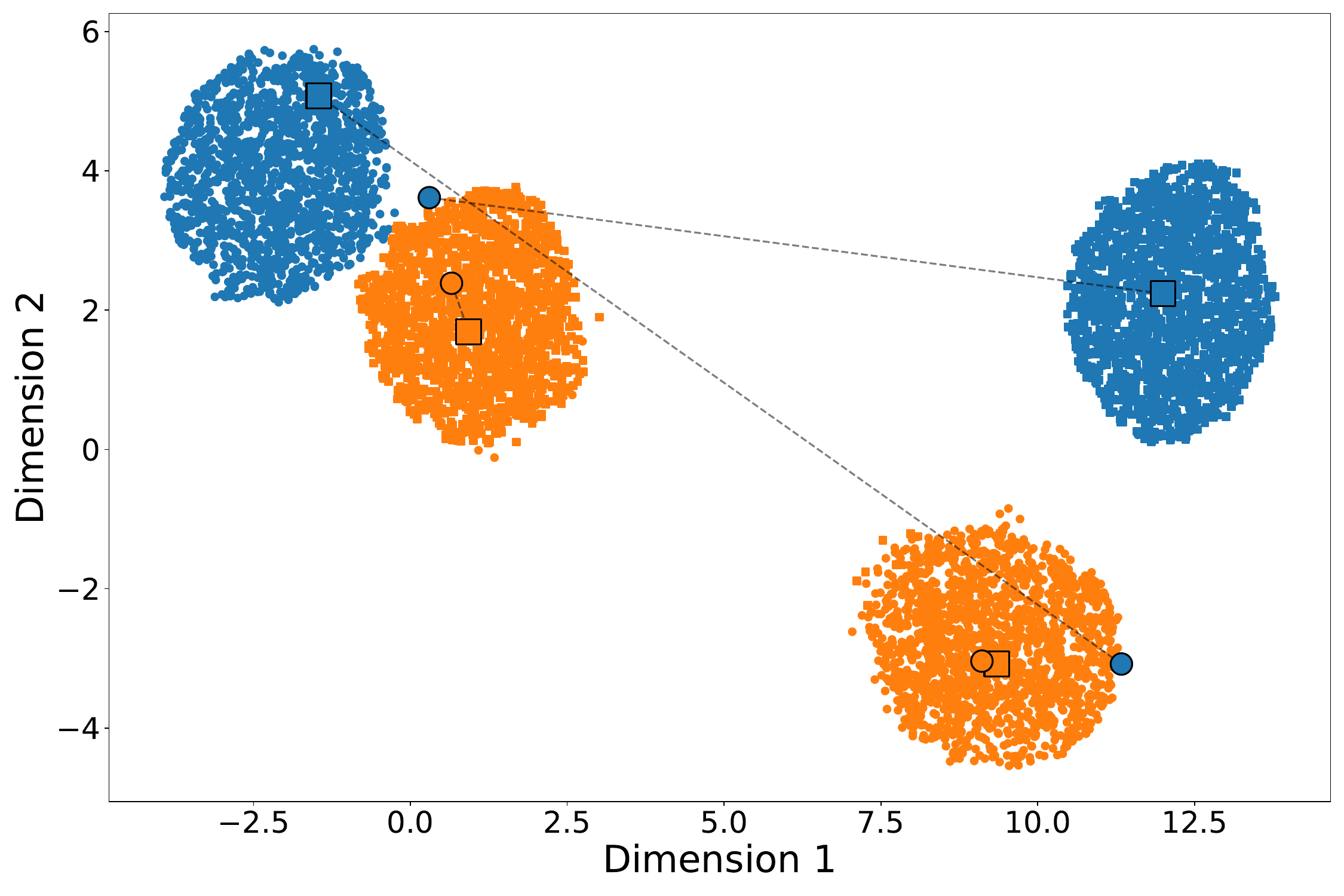}
        \subcaption{Without outliers}
        \label{fig:SAWSINE}
    \end{minipage}
    \begin{minipage}{0.4\textwidth}
        \centering
        \includegraphics[trim=0cm 0.5cm 0cm 0.2cm, clip, width=1\textwidth]{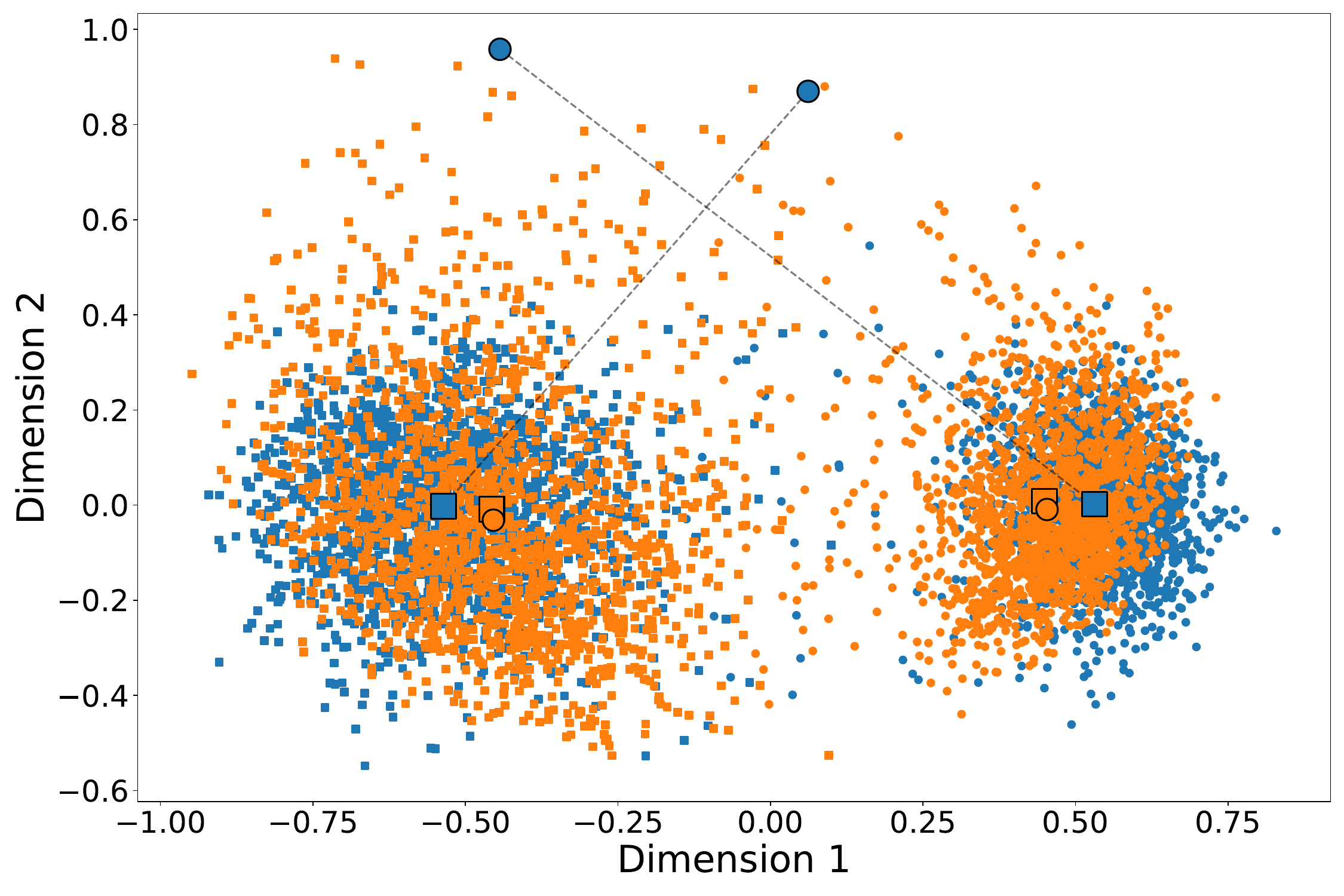}
        \subcaption{With Outliers}
        \label{fig:SAWSINE_outlier}
    \end{minipage}
    \hfill
    \begin{minipage}{0.19\textwidth}
    %\centering
        \includegraphics[trim=3.8cm -2.4cm 3.8cm 5.8cm, clip, width=1\textwidth]{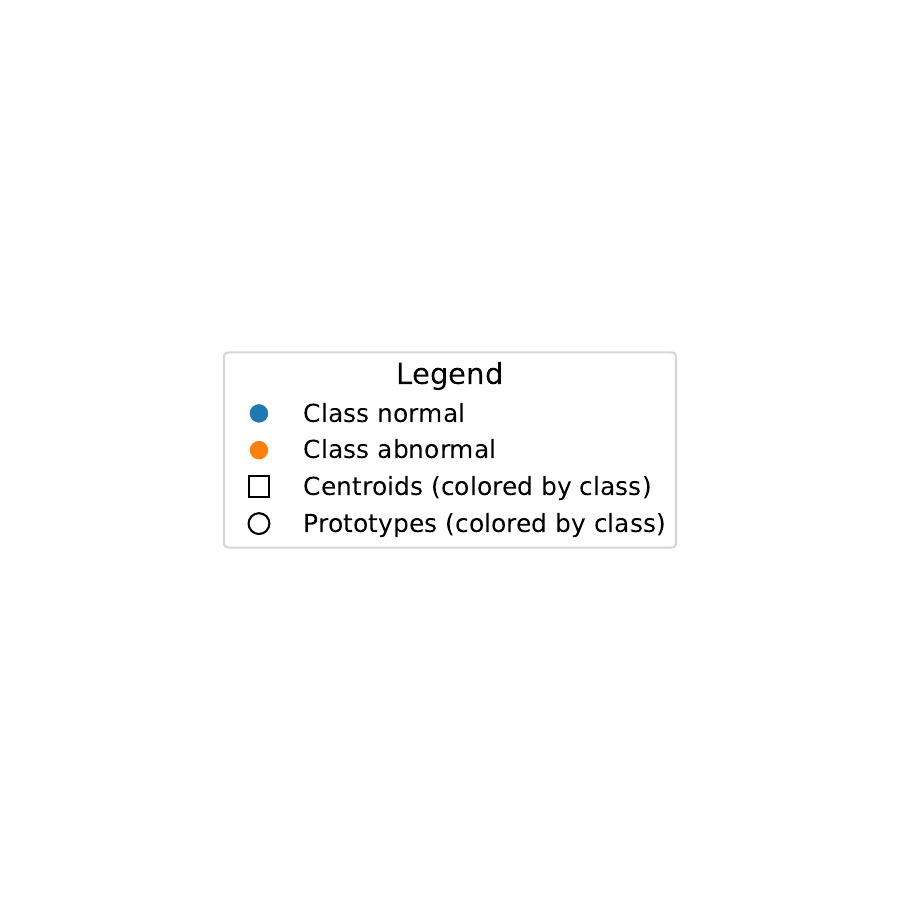}
    \end{minipage}
    \caption{Dimension-reduced representation of the latent space for the SAWSINE MAP model with index 5 evaluated with and without outliers. Due to the dimension reduction distances are distorted. Small dots represent input data, with lines connecting each prototype (circle) to its corresponding cluster centroid (square). Points are color-coded by class.}
    \label{fig:SAWSINE_outlier_latentspace}
\end{figure*}

Table~\ref{tab:results3} presents the results of the outlier analysis. Models for the SAWSINE dataset are evaluated with outliers incorporated either in the evaluation dataset alone or in both the training and evaluation phases. These models are compared to the original models for the unmodified dataset. Figure~\ref{fig:SAWSINE_outlier_latentspace} provides a visual comparison of exemplary latent spaces for the MAP model with index 5 evaluated with and without outliers.

\begin{table*}[t]
    \centering
    \caption{Benchmark results of the experiments for MAP and MSP for different datasets. The abbreviations in the column headings are the tested properties: (CR) Correctness, (CS) Consistency,
       (CN) Continuity, (CT) Contrastivity, (CC) Covariate Complexity, 
       (CP) Compactness, (CF) Confidence, (IC) Input Completeness, (CLS) Cohesion of Latent Space.}
    \label{tab:results2}
    \begin{tabular}{c| c c|c|ccccccccc|c}
        \toprule
        \textbf{Index} & \textbf{Dataset} & \textbf{Method} & \textbf{Val Loss} 
        & \textbf{CR} & \textbf{CS} & \textbf{CN} & \textbf{CT}
        & \textbf{CC} & \textbf{CP} & \textbf{CF} & \textbf{IC}
        & \textbf{CLS} & \textbf{Total}\\
        \midrule
        
1 & Wafer & MAP & MSE: $1.26$
              & 0.85 &    % CR
              0.28
                & 1.00 &  % CN
                  0.30 &  % CT
                  0.68 &  % CC
                  0.79 &  % CP
                  0.64 &  % CF
                  0.00    &  % IC
                  0.70 &  % CLS
                  0.58 \\% Total

2 & Wafer & MAP & MSE: $0.58$
              & 0.94 &
              0.46
                & 1.00 &
                  0.27 &
                  0.73 &
                  0.79 &
                  0.71 &
                  0.50 &
                  0.75 &
                  0.68 \\

3 & Wafer & MAP & MSE: $0.27$
              & 0.93 &
              0.73
                & 0.99 &
                  0.29 &
                  0.76 &
                  0.79 &
                  0.64 &
                  0.75 &
                  0.70 &
                  0.73 \\

4 & Wafer & MAP & MSE: $0.28$
              & 0.93 &
              0.79
                & 1.00 &
                  0.28 &
                  0.76 &
                  0.79 &
                  0.71 &
                  0.50 &
                  0.73 &
                  0.72 \\

\cellcolor{gray!20}1 & \cellcolor{gray!20}Wafer & \cellcolor{gray!20}MSP & \cellcolor{gray!20}MSE: $0.86$
            & \cellcolor{gray!20}1.00 & \cellcolor{gray!20}0.48
            & \cellcolor{gray!20}1.00 & \cellcolor{gray!20}0.43
            & \cellcolor{gray!20}0.54 & \cellcolor{gray!20}0.79
            & \cellcolor{gray!20}0.90 & \cellcolor{gray!20}0.00
            & \cellcolor{gray!20}0.54 & \cellcolor{gray!20}0.63 \\

\cellcolor{gray!20}2 & \cellcolor{gray!20}Wafer & \cellcolor{gray!20}MSP & \cellcolor{gray!20}MSE: $0.35$
            & \cellcolor{gray!20}0.71 & \cellcolor{gray!20}0.47
            & \cellcolor{gray!20}0.65 & \cellcolor{gray!20}0.27
            & \cellcolor{gray!20}0.66 & \cellcolor{gray!20}0.79
            & \cellcolor{gray!20}0.75 & \cellcolor{gray!20}0.40
            & \cellcolor{gray!20}0.76 & \cellcolor{gray!20}0.61 \\

\cellcolor{gray!20}3 & \cellcolor{gray!20}Wafer & \cellcolor{gray!20}MSP & \cellcolor{gray!20}MSE: $0.35$
            & \cellcolor{gray!20}0.87 & \cellcolor{gray!20}0.48
            & \cellcolor{gray!20}0.93 & \cellcolor{gray!20}0.36
            & \cellcolor{gray!20}0.72 & \cellcolor{gray!20}0.79
            & \cellcolor{gray!20}0.64 & \cellcolor{gray!20}0.20
            & \cellcolor{gray!20}0.83 & \cellcolor{gray!20}0.65 \\

\cellcolor{gray!20}4 & \cellcolor{gray!20}Wafer & \cellcolor{gray!20}MSP & \cellcolor{gray!20}MSE: $0.23$
            & \cellcolor{gray!20}0.41 & \cellcolor{gray!20}0.32
            & \cellcolor{gray!20}0.23 & \cellcolor{gray!20}0.27
            & \cellcolor{gray!20}0.82 & \cellcolor{gray!20}0.79
            & \cellcolor{gray!20}0.88 & \cellcolor{gray!20}0.75
            & \cellcolor{gray!20}0.78 & \cellcolor{gray!20}0.58 \\
   1 & SAWSINE & MAP & MSE: $0.37$
            & 0.74 & 0.76
            & 0.98 & 0.70 
            & 0.57 & 0.79 
            & 0.69 & 1.00 
            & 0.54 & 0.78 \\ 

2 & SAWSINE & MAP & MSE: $0.18$
            & 0.99 & 0.32
            & 0.99 & 0.41
            & 0.64 & 0.79
            & 0.52 & 0.75
            & 0.59 & 0.67 \\

3 & SAWSINE & MAP & MSE: $0.17$
            & 0.99 & 0.61
            & 0.99 & 0.32
            & 0.71 & 0.79
            & 0.55 & 1.00
            & 0.64 & 0.73 \\

4 & SAWSINE & MAP & MSE: $0.14$
            & 0.99 & 0.71
            & 0.98 & 0.36
            & 0.74 & 0.79
            & 0.53 & 0.75
            & 0.69 & 0.73 \\

5 & SAWSINE & MAP & MSE: $0.13$
            & 0.99 & 0.76
            & 1.00 & 0.30
            & 0.75 & 0.79
            & 0.57 & 1.00
            & 0.68 & 0.76 \\
    \cellcolor{gray!20}1 & \cellcolor{gray!20}SAWSINE & \cellcolor{gray!20}MSP & \cellcolor{gray!20}MSE: $0.18$
            & \cellcolor{gray!20}0.51 & \cellcolor{gray!20}0.16
            & \cellcolor{gray!20}0.94 & \cellcolor{gray!20}0.38
            & \cellcolor{gray!20}0.61 & \cellcolor{gray!20}0.79
            & \cellcolor{gray!20}0.36 & \cellcolor{gray!20}0.00
            & \cellcolor{gray!20}0.74 & \cellcolor{gray!20}0.50 \\

\cellcolor{gray!20}2 & \cellcolor{gray!20}SAWSINE & \cellcolor{gray!20}MSP & \cellcolor{gray!20}MSE: $0.14$
            & \cellcolor{gray!20}0.50 & \cellcolor{gray!20}0.44
            & \cellcolor{gray!20}0.95 & \cellcolor{gray!20}0.27
            & \cellcolor{gray!20}0.78 & \cellcolor{gray!20}0.79
            & \cellcolor{gray!20}0.81 & \cellcolor{gray!20}0.25
            & \cellcolor{gray!20}0.77 & \cellcolor{gray!20}0.62 \\

\cellcolor{gray!20}3 & \cellcolor{gray!20}SAWSINE & \cellcolor{gray!20}MSP & \cellcolor{gray!20}MSE: $0.14$
            & \cellcolor{gray!20}1.00 & \cellcolor{gray!20}0.43
            & \cellcolor{gray!20}0.94 & \cellcolor{gray!20}0.28
            & \cellcolor{gray!20}0.67 & \cellcolor{gray!20}0.79
            & \cellcolor{gray!20}0.84 & \cellcolor{gray!20}0.33
            & \cellcolor{gray!20}0.73 & \cellcolor{gray!20}0.67 \\

\cellcolor{gray!20}4 & \cellcolor{gray!20}SAWSINE & \cellcolor{gray!20}MSP & \cellcolor{gray!20}MSE: $0.14$
            & \cellcolor{gray!20}0.50 & \cellcolor{gray!20}0.30
            & \cellcolor{gray!20}0.87 & \cellcolor{gray!20}0.24
            & \cellcolor{gray!20}0.82 & \cellcolor{gray!20}0.79
            & \cellcolor{gray!20}0.92 & \cellcolor{gray!20}0.50
            & \cellcolor{gray!20}0.77 & \cellcolor{gray!20}0.63 \\  
1 & FordB & MAP & MSE: $1.02$
            & 0.54 & 0.34
            & 0.94 & 0.49
            & 0.51 & 0.79
            & 0.41 & 1.00
            & 0.50 & 0.61 \\

2 & FordB & MAP & MSE: $0.83$
            & 0.54 & 0.40
            & 0.96 & 0.53
            & 0.52 & 0.79
            & 0.40 & 0.75
            & 0.51 & 0.60 \\

3 & FordB & MAP & MSE: $0.62$
            & 0.54 & 0.40
            & 0.96 & 0.58
            & 0.51 & 0.79
            & 0.39 & 1.00
            & 0.51 & 0.63 \\

4 & FordB & MAP & MSE: $0.49$
            & 0.46 & 0.38
            & 0.96 & 0.58
            & 0.51 & 0.79
            & 0.39 & 0.75
            & 0.51 & 0.59 \\  
\cellcolor{gray!20}1 & \cellcolor{gray!20}FordB & \cellcolor{gray!20}MSP & \cellcolor{gray!20}MSE: $1.00$
            & \cellcolor{gray!20}0.49 & \cellcolor{gray!20}0.35
            & \cellcolor{gray!20}0.85 & \cellcolor{gray!20}0.34
            & \cellcolor{gray!20}0.65 & \cellcolor{gray!20}0.79
            & \cellcolor{gray!20}0.55 & \cellcolor{gray!20}0.00
            & \cellcolor{gray!20}0.67 & \cellcolor{gray!20}0.52 \\

\cellcolor{gray!20}2 & \cellcolor{gray!20}FordB & \cellcolor{gray!20}MSP & \cellcolor{gray!20}MSE: $1.00$
            & \cellcolor{gray!20}0.49 & \cellcolor{gray!20}0.42
            & \cellcolor{gray!20}0.61 & \cellcolor{gray!20}0.31
            & \cellcolor{gray!20}0.74 & \cellcolor{gray!20}0.79
            & \cellcolor{gray!20}0.77 & \cellcolor{gray!20}0.50
            & \cellcolor{gray!20}0.68 & \cellcolor{gray!20}0.59 \\

\cellcolor{gray!20}3 & \cellcolor{gray!20}FordB & \cellcolor{gray!20}MSP & \cellcolor{gray!20}MSE: $1.00$
            & \cellcolor{gray!20}0.47 & \cellcolor{gray!20}0.37
            & \cellcolor{gray!20}0.51 & \cellcolor{gray!20}0.39
            & \cellcolor{gray!20}0.56 & \cellcolor{gray!20}0.79
            & \cellcolor{gray!20}0.49 & \cellcolor{gray!20}0.00
            & \cellcolor{gray!20}0.74 & \cellcolor{gray!20}0.39 \\

\cellcolor{gray!20}4 & \cellcolor{gray!20}FordB & \cellcolor{gray!20}MSP & \cellcolor{gray!20}MSE: $1.00$
            & \cellcolor{gray!20}0.54 & \cellcolor{gray!20}0.29
            & \cellcolor{gray!20}0.52 & \cellcolor{gray!20}0.24
            & \cellcolor{gray!20}0.73 & \cellcolor{gray!20}0.79
            & \cellcolor{gray!20}0.73 & \cellcolor{gray!20}0.40
            & \cellcolor{gray!20}0.66 & \cellcolor{gray!20}0.54 \\
1 & FordA & MAP & MSE: $1.16$
            & 0.47 & 0.32
            & 0.95 & 0.44
            & 0.56 & 0.79
            & 0.42 & 1.00
            & 0.51 & 0.61 \\

2 & FordA & MAP & MSE: $1.05$
            & 0.47 & 0.34
            & 0.96 & 0.44
            & 0.54 & 0.79
            & 0.43 & 1.00
            & 0.51 & 0.61 \\

3 & FordA & MAP & MSE: $0.86$
            & 0.54 & 0.36
            & 0.96 & 0.50
            & 0.52 & 0.79
            & 0.41 & 0.75
            & 0.51 & 0.59 \\

4 & FordA & MAP & MSE: $0.70$
            & 0.47 & 0.37
            & 0.96 & 0.53
            & 0.51 & 0.79
            & 0.40 & 0.75
            & 0.51 & 0.59 \\

5 & FordA & MAP & MSE: $0.58$
            & 0.47 & 0.39
            & 0.97 & 0.56
            & 0.51 & 0.79
            & 0.39 & 0.75
            & 0.51 & 0.59 \\

\cellcolor{gray!20}1 & \cellcolor{gray!20}FordA & \cellcolor{gray!20}MSP & \cellcolor{gray!20}MSE: $1.00$
            & \cellcolor{gray!20}0.40 & \cellcolor{gray!20}0.54
            & \cellcolor{gray!20}0.56 & \cellcolor{gray!20}0.41
            & \cellcolor{gray!20}0.74 & \cellcolor{gray!20}0.79
            & \cellcolor{gray!20}0.75 & \cellcolor{gray!20}0.40
            & \cellcolor{gray!20}0.64 & \cellcolor{gray!20}0.58 \\

\cellcolor{gray!20}2 & \cellcolor{gray!20}FordA & \cellcolor{gray!20}MSP & \cellcolor{gray!20}MSE: $1.00$
            & \cellcolor{gray!20}0.49 & \cellcolor{gray!20}0.46
            & \cellcolor{gray!20}0.59 & \cellcolor{gray!20}0.29
            & \cellcolor{gray!20}0.76 & \cellcolor{gray!20}0.79
            & \cellcolor{gray!20}0.79 & \cellcolor{gray!20}0.25
            & \cellcolor{gray!20}0.62 & \cellcolor{gray!20}0.56 \\

\cellcolor{gray!20}3 & \cellcolor{gray!20}FordA & \cellcolor{gray!20}MSP & \cellcolor{gray!20}MSE: $1.00$
            & \cellcolor{gray!20}0.58 & \cellcolor{gray!20}0.31
            & \cellcolor{gray!20}0.61 & \cellcolor{gray!20}0.31
            & \cellcolor{gray!20}0.72 & \cellcolor{gray!20}0.79
            & \cellcolor{gray!20}0.75 & \cellcolor{gray!20}0.50
            & \cellcolor{gray!20}0.67 & \cellcolor{gray!20}0.58 \\

\cellcolor{gray!20}4 & \cellcolor{gray!20}FordA & \cellcolor{gray!20}MSP & \cellcolor{gray!20}MSE: $0.99$
            & \cellcolor{gray!20}0.36& \cellcolor{gray!20}0.32
            & \cellcolor{gray!20}0.50 & \cellcolor{gray!20}0.24
            & \cellcolor{gray!20}0.71 & \cellcolor{gray!20}0.79
            & \cellcolor{gray!20}0.75 & \cellcolor{gray!20}0.33
            & \cellcolor{gray!20}0.67 & \cellcolor{gray!20}0.52 \\

1 & ECG200 & MAP & MSE: $1.45$
            & 0.69 & 0.28
            & 0.98 & 0.43
            & 0.68 & 0.79
            & 0.67 & 0.67
            & 0.63 & 0.65 \\

2 & ECG200 & MAP & MSE: $0.70$
            & 0.78 & 0.34
            & 0.98 & 0.37
            & 0.66 & 0.79
            & 0.68 & 0.80
            & 0.64 & 0.67 \\
            
%2 & ECG200 & MAP & MSE: $0.60$
%            & 0.85 & 0.35
%            & 0.98 & 0.37
%            & 0.67 & 0.79
%            & 0.63 & 0.67
%            & 0.64 & 0.66 \\

3 & ECG200 & MAP & MSE: $0.55$
            & 0.85 & 0.40
            & 1.00 & 0.41
            & 0.65 & 0.79
            & 0.69 & 0.67
            & 0.66 & 0.68 \\
            
4 & ECG200 & MAP & MSE: $0.38$
            & 0.88 & 0.46
            & 1.00 & 0.30
            & 0.69 & 0.79
            & 0.61 & 0.57
            & 0.66 & 0.66 \\

5 & ECG200 & MAP & MSE: $0.26$
            & 0.88 & 0.46
            & 0.97 & 0.30
            & 0.69 & 0.79
            & 0.60 & 0.57
            & 0.64 & 0.66 \\ 
            
\cellcolor{gray!20}1 & \cellcolor{gray!20} ECG200 &\cellcolor{gray!20}MSP & \cellcolor{gray!20}MSE: $1.04$
            & \cellcolor{gray!20}1.00 & \cellcolor{gray!20}0.37
            &\cellcolor{gray!20}1.00 &\cellcolor{gray!20}0.49
            &\cellcolor{gray!20}0.50 &\cellcolor{gray!20}0.79
            &\cellcolor{gray!20}0.55 &\cellcolor{gray!20}0.00
            &\cellcolor{gray!20}0.60 &\cellcolor{gray!20}0.59\\

\cellcolor{gray!20}2 & \cellcolor{gray!20}ECG200 & \cellcolor{gray!20}MSP & \cellcolor{gray!20}MSE: $0.78$
            & \cellcolor{gray!20}1.00 & \cellcolor{gray!20}0.45
            & \cellcolor{gray!20}1.00 & \cellcolor{gray!20}0.46
            & \cellcolor{gray!20}0.50 & \cellcolor{gray!20}0.79
            & \cellcolor{gray!20}0.43 & \cellcolor{gray!20}0.00
            & \cellcolor{gray!20}0.52 & \cellcolor{gray!20}0.57 \\

\cellcolor{gray!20}3 & \cellcolor{gray!20}ECG200 & \cellcolor{gray!20}MSP & \cellcolor{gray!20}MSE: $0.36$
            & \cellcolor{gray!20}1.00 & \cellcolor{gray!20}0.55
            & \cellcolor{gray!20}0.95 & \cellcolor{gray!20}0.37
            & \cellcolor{gray!20}0.53 & \cellcolor{gray!20}0.79
            & \cellcolor{gray!20}0.47 & \cellcolor{gray!20}0.00
            & \cellcolor{gray!20}0.39 & \cellcolor{gray!20}0.56 \\

%\cellcolor{gray!20}4 & \cellcolor{gray!20}ECG200 & \cellcolor{gray!20}MSP & \cellcolor{gray!20}MSE: $0.30$
%            & \cellcolor{gray!20}0.34 & \cellcolor{gray!20}0.60
%            & \cellcolor{gray!20}0.90 & \cellcolor{gray!20}0.43
%            & \cellcolor{gray!20}0.57 & \cellcolor{gray!20}0.79
 %           & \cellcolor{gray!20}0.62 & \cellcolor{gray!20}0.80
 %           & \cellcolor{gray!20}0.64 & \cellcolor{gray!20}0.63 \\
\cellcolor{gray!20}4 & \cellcolor{gray!20}ECG200 & \cellcolor{gray!20}MSP & \cellcolor{gray!20}MSE: $0.30$
            & \cellcolor{gray!20}0.86 & \cellcolor{gray!20}0.60
            & \cellcolor{gray!20}0.86 & \cellcolor{gray!20}0.33
            & \cellcolor{gray!20}0.74 & \cellcolor{gray!20}0.79
            & \cellcolor{gray!20}0.66 & \cellcolor{gray!20}0.20
            & \cellcolor{gray!20}0.73 & \cellcolor{gray!20}0.64 \\
            
\cellcolor{gray!20}5 & \cellcolor{gray!20}ECG200 & \cellcolor{gray!20}MSP & \cellcolor{gray!20}MSE: $0.29$
            & \cellcolor{gray!20}0.96 & \cellcolor{gray!20}0.60
            & \cellcolor{gray!20}0.83 & \cellcolor{gray!20}0.38
            & \cellcolor{gray!20}0.61 & \cellcolor{gray!20}0.79
            & \cellcolor{gray!20}0.57 & \cellcolor{gray!20}0.00
            & \cellcolor{gray!20}0.73 & \cellcolor{gray!20}0.61 \\
                  
1 & StarLightCurve & MAP & MSE: $1.07$
            & 0.82 & 0.26
            & 0.98 & 0.32
            & 0.68 & 0.67
            & 0.76 & 0.67
            & 0.58 & 0.64 \\

2 & StarLightCurve & MAP & MSE: $0.76$
            & 0.85 & 0.26
            & 0.93 & 0.34
            & 0.64 & 0.67
            & 0.61 & 0.71
            & 0.57 & 0.62 \\

3 & StarLightCurve & MAP & MSE: $0.55$
            & 0.85 & 0.32
            & 0.94 & 0.32
            & 0.54 & 0.79
            & 0.63 & 0.63
            & 0.55 & 0.62 \\
            
4 & StarLightCurve & MAP & MSE: $0.31$
            & 0.85 & 0.54
            & 0.95 & 0.34
            & 0.61 & 0.67
            & 0.65 & 0.57
            & 0.56 & 0.64 \\

5 & StarLightCurve & MAP & MSE: $0.13$
            & 0.85 & 0.65
            & 0.97 & 0.31
            & 0.65 & 0.67
            & 0.63 & 1.00
            & 0.57 & 0.71 \\ 
            
\cellcolor{gray!20}1 & \cellcolor{gray!20} StarLightCurve &\cellcolor{gray!20}MSP & \cellcolor{gray!20}MSE: $0.42$
            & \cellcolor{gray!20}0.21 & \cellcolor{gray!20}0.38
            &\cellcolor{gray!20}0.97 &\cellcolor{gray!20}0.33
            &\cellcolor{gray!20}0.54 &\cellcolor{gray!20}0.67
            &\cellcolor{gray!20}0.36 &\cellcolor{gray!20}0.00
            &\cellcolor{gray!20}0.58 &\cellcolor{gray!20}0.45\\

\cellcolor{gray!20}2 & \cellcolor{gray!20}StarLightCurve & \cellcolor{gray!20}MSP & \cellcolor{gray!20}MSE: $0.28$
            & \cellcolor{gray!20}0.29 & \cellcolor{gray!20}0.34
            & \cellcolor{gray!20}0.99 & \cellcolor{gray!20}0.32
            & \cellcolor{gray!20}0.55 & \cellcolor{gray!20}0.67
            & \cellcolor{gray!20}0.59 & \cellcolor{gray!20}0.00
            & \cellcolor{gray!20}0.65 & \cellcolor{gray!20}0.49 \\

\cellcolor{gray!20}3 & \cellcolor{gray!20}StarLightCurve & \cellcolor{gray!20}MSP & \cellcolor{gray!20}MSE: $0.22$
            & \cellcolor{gray!20}0.38 & \cellcolor{gray!20}0.56
            & \cellcolor{gray!20}0.60 & \cellcolor{gray!20}0.35
            & \cellcolor{gray!20}0.59 & \cellcolor{gray!20}0.67
            & \cellcolor{gray!20}0.66 & \cellcolor{gray!20}0.00
            & \cellcolor{gray!20}0.66 & \cellcolor{gray!20}0.50 \\

\cellcolor{gray!20}4 & \cellcolor{gray!20}StarLightCurve & \cellcolor{gray!20}MSP & \cellcolor{gray!20}MSE: $0.20$
            & \cellcolor{gray!20}0.39 & \cellcolor{gray!20}0.42
            & \cellcolor{gray!20}0.25 & \cellcolor{gray!20}0.24
            & \cellcolor{gray!20}0.58 & \cellcolor{gray!20}0.67
            & \cellcolor{gray!20}0.62 & \cellcolor{gray!20}0.11
            & \cellcolor{gray!20}0.66 & \cellcolor{gray!20}0.44 \\
            
\cellcolor{gray!20}5 & \cellcolor{gray!20}StarLightCurve & \cellcolor{gray!20}MSP & \cellcolor{gray!20}MSE: $0.12$
            & \cellcolor{gray!20}0.40 & \cellcolor{gray!20}0.62
            & \cellcolor{gray!20}0.72 & \cellcolor{gray!20}0.27
            & \cellcolor{gray!20}0.80 & \cellcolor{gray!20}0.67
            & \cellcolor{gray!20}0.76 & \cellcolor{gray!20}0.38
            & \cellcolor{gray!20}0.74 & \cellcolor{gray!20}0.59 \\
        \bottomrule
    \end{tabular}
\end{table*}

\begin{table*}[t]
    \centering
    \caption{Benchmark results of the outlier analysis for MAP and MSP for SAWSINE datasets. Here, {\textquotedblleft{mixed}\textquotedblright} refers to the original models evaluated on the outlier dataset, while {\textquotedblleft{Outlier}\textquotedblright} denotes the models that was both trained and evaluated using the outlier dataset. The abbreviations in the column headings are the tested properties: (CR) Correctness, (CS) Consistency,
       (CN) Continuity, (CT) Contrastivity, (CC) Covariate Complexity, 
       (CP) Compactness, (CF) Confidence, (IC) Input Completeness, (CLS) Cohesion of Latent Space.}
    \label{tab:results3}
    \begin{tabular}{c| c c|c|ccccccccc|c}
        \toprule
        \textbf{Index} & \textbf{Dataset} & \textbf{Method} & \textbf{Val Loss} 
        & \textbf{CR} & \textbf{CS} & \textbf{CN} & \textbf{CT}
        & \textbf{CC} & \textbf{CP} & \textbf{CF} & \textbf{IC}
        & \textbf{CLS} & \textbf{Total}\\
        \midrule
1 & SAWSINE & MAP & MSE: $0.37$
            & 0.74 & 0.76
            & 0.98 & 0.70 
            & 0.57 & 0.79 
            & 0.69 & 1.00 
            & 0.54 & 0.78 \\ 

2 & SAWSINE & MAP & MSE: $0.18$
            & 0.99 & 0.32
            & 0.99 & 0.41
            & 0.64 & 0.79
            & 0.52 & 0.75
            & 0.59 & 0.67 \\

3 & SAWSINE & MAP & MSE: $0.17$
            & 0.99 & 0.61
            & 0.99 & 0.32
            & 0.71 & 0.79
            & 0.55 & 1.00
            & 0.64 & 0.73 \\

4 & SAWSINE & MAP & MSE: $0.14$
            & 0.99 & 0.71
            & 0.98 & 0.36
            & 0.74 & 0.79
            & 0.53 & 0.75
            & 0.69 & 0.73 \\

5 & SAWSINE & MAP & MSE: $0.13$
            & 0.99 & 0.76
            & 1.00 & 0.30
            & 0.75 & 0.79
            & 0.57 & 1.00
            & 0.68 & 0.76 \\
    \cellcolor{gray!20}1 & \cellcolor{gray!20}SAWSINE & \cellcolor{gray!20}MSP & \cellcolor{gray!20}MSE: $0.18$
            & \cellcolor{gray!20}0.51 & \cellcolor{gray!20}0.16
            & \cellcolor{gray!20}0.94 & \cellcolor{gray!20}0.38
            & \cellcolor{gray!20}0.61 & \cellcolor{gray!20}0.79
            & \cellcolor{gray!20}0.36 & \cellcolor{gray!20}0.00
            & \cellcolor{gray!20}0.74 & \cellcolor{gray!20}0.50 \\

\cellcolor{gray!20}2 & \cellcolor{gray!20}SAWSINE & \cellcolor{gray!20}MSP & \cellcolor{gray!20}MSE: $0.14$
            & \cellcolor{gray!20}0.50 & \cellcolor{gray!20}0.44
            & \cellcolor{gray!20}0.95 & \cellcolor{gray!20}0.27
            & \cellcolor{gray!20}0.78 & \cellcolor{gray!20}0.79
            & \cellcolor{gray!20}0.81 & \cellcolor{gray!20}0.25
            & \cellcolor{gray!20}0.77 & \cellcolor{gray!20}0.62 \\

\cellcolor{gray!20}3 & \cellcolor{gray!20}SAWSINE & \cellcolor{gray!20}MSP & \cellcolor{gray!20}MSE: $0.14$
            & \cellcolor{gray!20}1.00 & \cellcolor{gray!20}0.43
            & \cellcolor{gray!20}0.94 & \cellcolor{gray!20}0.28
            & \cellcolor{gray!20}0.67 & \cellcolor{gray!20}0.79
            & \cellcolor{gray!20}0.84 & \cellcolor{gray!20}0.33
            & \cellcolor{gray!20}0.73 & \cellcolor{gray!20}0.67 \\

\cellcolor{gray!20}4 & \cellcolor{gray!20}SAWSINE & \cellcolor{gray!20}MSP & \cellcolor{gray!20}MSE: $0.14$
            & \cellcolor{gray!20}0.50 & \cellcolor{gray!20}0.30
            & \cellcolor{gray!20}0.87 & \cellcolor{gray!20}0.24
            & \cellcolor{gray!20}0.82 & \cellcolor{gray!20}0.79
            & \cellcolor{gray!20}0.92 & \cellcolor{gray!20}0.50
            & \cellcolor{gray!20}0.77 & \cellcolor{gray!20}0.63 \\  
            
1 & SAWSINE mixed & MAP & MSE: $0.42$
            & 0.71 & 0.76
            & 0.98 & 0.70 
            & 0.57 & 0.79 
            & 0.69 & 1.00 
            & 0.54 & 0.75 \\ 

2 & SAWSINE mixed & MAP & MSE: $0.22$
            & 0.97 & 0.32
            & 0.98 & 0.41
            & 0.64 & 0.79
            & 0.51 & 0.75
            & 0.58 & 0.70 \\

3 & SAWSINE mixed & MAP & MSE: $0.21$
            & 0.98 & 0.61
            & 0.99 & 0.32
            & 0.71 & 0.79
            & 0.55 & 1.00
            & 0.64 & 0.74 \\

4 & SAWSINE mixed & MAP & MSE: $0.18$
            & 0.97 & 0.71
            & 0.98 & 0.36
            & 0.74 & 0.79
            & 0.53 & 0.75
            & 0.68 & 0.72 \\

5 & SAWSINE mixed & MAP & MSE: $0.17$
            & 0.98 & 0.76
            & 0.99 & 0.30
            & 0.75 & 0.79
            & 0.56 & 1.00
            & 0.67 & 0.75 \\
\cellcolor{gray!20}1 & \cellcolor{gray!20}SAWSINE mixed & \cellcolor{gray!20}MSP & \cellcolor{gray!20}MSE: $0.22$
            & \cellcolor{gray!20}0.50 & \cellcolor{gray!20}0.16
            & \cellcolor{gray!20}0.93 & \cellcolor{gray!20}0.38
            & \cellcolor{gray!20}0.68 & \cellcolor{gray!20}0.79
            & \cellcolor{gray!20}0.33 & \cellcolor{gray!20}0.00
            & \cellcolor{gray!20}0.75 & \cellcolor{gray!20}0.54 \\

\cellcolor{gray!20}2 & \cellcolor{gray!20}SAWSINE mixed & \cellcolor{gray!20}MSP & \cellcolor{gray!20}MSE: $0.21$
            & \cellcolor{gray!20}0.50 & \cellcolor{gray!20}0.44
            & \cellcolor{gray!20}0.95 & \cellcolor{gray!20}0.27
            & \cellcolor{gray!20}0.74 & \cellcolor{gray!20}0.79
            & \cellcolor{gray!20}0.79 & \cellcolor{gray!20}0.25
            & \cellcolor{gray!20}0.77 & \cellcolor{gray!20}0.63 \\

\cellcolor{gray!20}3 & \cellcolor{gray!20}SAWSINE mixed & \cellcolor{gray!20}MSP & \cellcolor{gray!20}MSE: $0.19$
            & \cellcolor{gray!20}0.99 & \cellcolor{gray!20}0.43
            & \cellcolor{gray!20}0.93 & \cellcolor{gray!20}0.28
            & \cellcolor{gray!20}0.74 & \cellcolor{gray!20}0.79
            & \cellcolor{gray!20}0.81 & \cellcolor{gray!20}0.11
            & \cellcolor{gray!20}0.76 & \cellcolor{gray!20}0.68 \\

\cellcolor{gray!20}4 & \cellcolor{gray!20}SAWSINE mixed & \cellcolor{gray!20}MSP & \cellcolor{gray!20}MSE: $0.19$
            & \cellcolor{gray!20}0.49 & \cellcolor{gray!20}0.30
            & \cellcolor{gray!20}0.85 & \cellcolor{gray!20}0.24
            & \cellcolor{gray!20}0.83 & \cellcolor{gray!20}0.79
            & \cellcolor{gray!20}0.90 & \cellcolor{gray!20}0.50
            & \cellcolor{gray!20}0.81 & \cellcolor{gray!20}0.68 \\
            
   1 & SAWSINE Outlier & MAP & MSE: $0.62$
            & 0.92 & 0.36
            & 0.97 & 0.58 
            & 0.59 & 0.79 
            & 0.57 & 1.00 
            & 0.54 & 0.74 \\ 

2 & SAWSINE Outlier & MAP & MSE: $0.35$
            & 0.96 & 0.45
            & 0.98 & 0.54
            & 0.61 & 0.79
            & 0.52 & 0.75
            & 0.55 & 0.71 \\

3 & SAWSINE Outlier & MAP & MSE: $0.30$
            & 0.98 & 0.42
            & 0.98 & 0.45
            & 0.61 & 0.79
            & 0.50 & 1.00
            & 0.56 & 0.74 \\

4 & SAWSINE Outlier & MAP & MSE: $0.27$
            & 0.98 & 0.44
            & 0.98 & 0.40
            & 0.63 & 0.79
            & 0.47 & 1.00
            & 0.57 & 0.73 \\

5 & SAWSINE Outlier & MAP & MSE: $0.24$
            & 0.98 & 0.50
            & 0.99 & 0.42
            & 0.65 & 0.79
            & 0.47 & 1.00
            & 0.60 & 0.70 \\
%\cellcolor{gray!20}1 & \cellcolor{gray!20}SAWSINE Outlier & \cellcolor{gray!20}MSP & \cellcolor{gray!20}MSE: $0.24$
%            & \cellcolor{gray!20}0.51 & \cellcolor{gray!20}0.49
%            & \cellcolor{gray!20}0.97 & \cellcolor{gray!20}0.50
%            & \cellcolor{gray!20}0.51 & \cellcolor{gray!20}0.79
%            & \cellcolor{gray!20}0.54 & \cellcolor{gray!20}0.00
%            & \cellcolor{gray!20}0.69 & \cellcolor{gray!20}0.56 \\

\cellcolor{gray!20}1 & \cellcolor{gray!20}SAWSINE Outlier & \cellcolor{gray!20}MSP & \cellcolor{gray!20}MSE: $0.22$
            & \cellcolor{gray!20}0.98 & \cellcolor{gray!20}0.54
            & \cellcolor{gray!20}0.98 & \cellcolor{gray!20}0.37
            & \cellcolor{gray!20}0.54 & \cellcolor{gray!20}0.79
            & \cellcolor{gray!20}0.52 & \cellcolor{gray!20}0.00
            & \cellcolor{gray!20}0.75 & \cellcolor{gray!20}0.62 \\

\cellcolor{gray!20}2 & \cellcolor{gray!20}SAWSINE Outlier & \cellcolor{gray!20}MSP & \cellcolor{gray!20}MSE: $0.21$
            & \cellcolor{gray!20}0.84 & \cellcolor{gray!20}0.65
            & \cellcolor{gray!20}0.94 & \cellcolor{gray!20}0.38
            & \cellcolor{gray!20}0.64 & \cellcolor{gray!20}0.79
            & \cellcolor{gray!20}0.67 & \cellcolor{gray!20}0.20
            & \cellcolor{gray!20}0.76 & \cellcolor{gray!20}0.65 \\

\cellcolor{gray!20}3 & \cellcolor{gray!20}SAWSINE Outlier & \cellcolor{gray!20}MSP & \cellcolor{gray!20}MSE: $0.19$
            & \cellcolor{gray!20}0.57 & \cellcolor{gray!20}0.21
            & \cellcolor{gray!20}0.98 & \cellcolor{gray!20}0.33
            & \cellcolor{gray!20}0.68 & \cellcolor{gray!20}0.79
            & \cellcolor{gray!20}0.72 & \cellcolor{gray!20}0.25
            & \cellcolor{gray!20}0.77 & \cellcolor{gray!20}0.63 \\  
            
\cellcolor{gray!20}4 & \cellcolor{gray!20}SAWSINE Outlier & \cellcolor{gray!20}MSP & \cellcolor{gray!20}MSE: $0.19$
            & \cellcolor{gray!20}0.53 & \cellcolor{gray!20}0.64
            & \cellcolor{gray!20}0.96 & \cellcolor{gray!20}0.29
            & \cellcolor{gray!20}0.77 & \cellcolor{gray!20}0.79
            & \cellcolor{gray!20}0.82 & \cellcolor{gray!20}0.50
            & \cellcolor{gray!20}0.80 & \cellcolor{gray!20}0.68 \\  
        \bottomrule
    \end{tabular}
\end{table*}

\end{document}